\documentclass[11pt]{article}

\usepackage[preprint]{acl}

\usepackage{times}
\usepackage{latexsym}

\usepackage[T1]{fontenc}

\usepackage[utf8]{inputenc}

\usepackage{microtype}

\usepackage{inconsolata}

\usepackage{graphicx}
\usepackage{amsmath,amssymb,amsfonts}%
\usepackage{amsthm}%
\usepackage{mathrsfs}%
\usepackage[title]{appendix}%
\usepackage{xcolor}%
\usepackage{textcomp}%
\usepackage{manyfoot}%
\usepackage{booktabs}%
\usepackage{algorithm}%
\usepackage{algorithmicx}%
\usepackage{algpseudocode}%
\usepackage{listings}%

\usepackage{nicefrac}       
\usepackage{tabularx}
\usepackage{multicol}
\usepackage{multirow}
\usepackage{balance}
\usepackage{subcaption}

\title{Automatic Combination of Sample Selection Strategies for Few-Shot Learning}

\author{Branislav Pecher$^\spadesuit$$^\dagger$, Ivan Srba$^\dagger$, Maria Bielikova$^\dagger$, Joaquin Vanschoren$^\ddagger$  \\
$^\spadesuit$ Faculty of Information Technology, Brno University of Technology, Brno, Czechia\\
$^\dagger$ Kempelen Institute of Intelligent Technologies, Bratislava, Slovakia\\
$^\ddagger$ Eindhoven University of Technology, Eindhoven, Netherlands\\
\texttt{\{branislav.pecher, ivan.srba, maria.bielikova\}@kinit.sk, j.vanschoren@tue.nl}}

\begin{document}
\maketitle
\begin{abstract}
In few-shot learning, the selection of samples has a significant impact on the performance of the model. While effective sample selection strategies are well-established in supervised settings, research on large language models largely overlooks them, favouring strategies specifically tailored to individual in-context learning settings. In this paper, we propose a new method for Automatic Combination of SamplE Selection Strategies (ACSESS) to leverage the strengths and complementarity of various well-established selection objectives. We investigate and compare the impact of 23 sample selection strategies on the performance of 5 in-context learning models and 3 few-shot learning approaches (meta-learning, few-shot fine-tuning) over 6 text and 8 image datasets. The experimental results show that the combination of strategies through the ACSESS method consistently outperforms all individual selection strategies and performs on par or exceeds the in-context learning specific baselines. Lastly, we demonstrate that sample selection remains effective even on smaller datasets, yielding the greatest benefits when only a few shots are selected, while its advantage diminishes as the number of shots increases.
\end{abstract}

\section{Introduction}

Many domains are characterised by a labelled data scarcity due to data collection/annotation costs or privacy considerations, making the training of typical deep learning models unfeasible. Few-shot learning addresses the challenge of adapting models to new tasks when only a handful of labelled samples are available~\citep{song2023survey}. Two main approaches exist: 1) \textit{in-context learning}, where a pretrained large language model is conditioned on few examples without any parameter updates~\citep{dong2022survey, liu2023pre}; and 2) \textit{gradient-based few-shot learning}, which includes meta-learning and fine-tuning, where the knowledge from relevant datasets is transferred to a new problem using gradient-based training~\citep{song2023survey, chen2019closer, vanschoren2018meta, hospedales2021meta}. More details regarding their differences are included in Appendix~\ref{app:fsl-vs-icl}.

The choice of which sample to use is critical in both approaches, as performance can vary drastically depending on sample quality. Existing studies in different settings, such as in LLM alignment ~\citep{zhou2024lima}, have shown that \textbf{curating a smaller subset of high-quality samples can lead to better performance than training on the full set of available samples}. This holds for few-shot learning as well, especially for noisy or unbalanced datasets or for in-context learning, which is known to be sensitive to sample choice~\citep{pecher2023effects, agarwal2021sensitivity, koksal2022meal}.

Many works have already proposed diverse sample selection strategies to identify an optimal set of samples. Such a \textit{sample selection strategy} follows its specific \textit{objective} by considering one or few \textit{properties} of available samples, e.g., to maximise similarity, diversity, informativeness or quality of selected samples~\citep{li-qiu-2023-finding, zhang-etal-2022-active, chang-jia-2023-data}. With the popularisation of large language models (LLMs), many new selection strategies are proposed for in-context learning. However, they are often tailored only for specific settings, limiting their applicability. At the same time, the well-established and long-standing selection strategies that were shown to perform well in supervised settings, such as active learning, are mostly ignored as part of LLM sample selection~\citep{pecher2023effects, albalak2024a}.

In this work, we \textbf{demonstrate that well-established strategies can match or exceed the performance of in-context learning specific strategies}. To achieve this, we \textbf{automatically and optimally combine the sample selection objectives used by well-established sample selection strategies to identify samples with complementary properties}. In order to leverage the strengths of different sample selection strategies, we propose \textbf{ACSESS}, an effective method for \textbf{A}utomatic \textbf{C}ombination of \textbf{S}ampl\textbf{E} \textbf{S}election \textbf{S}trategies. First, a subset of relevant strategies that can improve the overall success of few-shot learning is identified. Afterwards, the selection objectives from the identified strategies are combined together, using weighting based on their expected contribution, in order to identify the most informative and high-quality samples that can provide the most benefit. Our contributions and findings are as follows\footnote{To support replicability and extension of our results, we openly publish the source code of our experiments at \url{https://github.com/kinit-sk/ACSESS}}:
\begin{itemize}
    \item We provide the first large-scale study of 23 sample selection strategies across 5 in-context learning models and 3 few-shot learning approaches, including meta-learning and few-shot fine-tuning, across 6 text and 8 image datasets. We show that sample selection can yield consistent gains (up to $5$  and $3$ percentage points for in-context learning and gradient few-shot learning, respectively), though the effects depend heavily on dataset and approach. 
    \item We propose \textbf{ACSESS}, a method that automatically combines and weights sample selection objectives to leverage their strengths and identify samples with complementary properties (i.e., informativeness, representativeness and learnability). Our results show that the combination of well-established strategies identified by the ACSESS method consistently leads to performance on par or better than the existing in-context learning specific baselines.  
    \item Through ablation studies, we find following key insights: 1) sample selection has higher impact when the number of shots is low or when using noisy datasets; 2) at higher number of shots (30-40 on average), the impact of sample selection is negligible as all strategies regress to random selection; 3) after a certain point, the boost in performance from using more shots becomes negligible (30 shots for in-context learning; 50-shot for gradient few-shot learning); 4) sample selection is beneficial even for small dataset sizes (achieving similar performance when selecting from only 25\% for in-context learning or 10\% of the dataset for gradient few-shot learning).
\end{itemize}
\section{Related Work}

A large body of work is dedicated to selecting high-quality samples for in-context learning, where the overall performance was found to be sensitive to this choice, leading to large variability in results~\citep{pecher2023effects, koksal2022meal, zhang-etal-2022-active}. In the supervised setting, the selection is mostly focused on distilling datasets to a lower number of samples~\citep{yu2023dataset}, selecting a core set representative of the full set~\citep{guo2022deepcore}, or reducing annotation costs using active learning~\citep{ren2021survey}. However, the impact of these strategies remains underexplored for other few-shot learning approaches.

Within the majority of the sample selection strategies, the sample selection objective takes into consideration \textbf{a single property} of available samples. To this end, various heuristics and unsupervised metrics serve as an estimate of their potential to increase the model's performance. For in-context learning, the most popular approach is selecting samples based on the \textbf{similarity} to the test sample, differing only in employed similarity measure and representation~\citep{liu-etal-2022-makes, an-etal-2023-skill, zemlyanskiy-etal-2022-generate, gupta-etal-2022-retronlu, pasupat-etal-2021-controllable, wang-etal-2022-training, nashid2023retrieval, agrawal-etal-2023-context, gao-etal-2021-making}. Besides similarity, the samples are often selected based on their \textbf{informativeness} or the \textbf{uncertainty}, often using active learning strategies ~\citep{koksal2022meal, schroder-etal-2022-revisiting, margatina-etal-2021-active, park2022active, mavromatis2023examples}, bias of the samples~\citep{ma2023fairness} or adversarial training~\citep{agarwal2021sensitivity}. Other methods define a notion of \textbf{quality} for each sample, either by prompting a LLM to rate the samples~\citep{shin-etal-2021-constrained} or observing how the inclusion or removal of the sample affects the performance~\citep{maronikolakis-etal-2023-improving, nguyen2023context}. When considering \textbf{representativeness}, the samples are selected based on how well they are representative of the full dataset~\citep{guo2022deepcore, killamsetty2021glister, mirzasoleiman2020coresets, paul2021grand, coleman2020Selection}. \textbf{Learnability} metric either determines how easy it is to learn the specific sample~\citep{swayamdipta-etal-2020-dataset, zhang-plank-2021-cartography-active}, or how often the sample is forgotten after being learned~\citep{toneva2018empirical}.

Finally, some approaches \textbf{balance multiple sample properties} at the same time. Often, the similarity of samples is combined with their diversity~\citep{wu2023selfadaptive, qin2023context} or the informativeness of the samples is balanced with their representativeness~\citep{su2022selective, levy-etal-2023-diverse, ye-etal-2023-complementary, liu2023towards}. In specific cases, a two step sample search is done, such as finding set of informative samples and then using diversity guided search to improve this set~\citep{li-qiu-2023-finding}, or using different uncertainty measures to select the top samples~\citep{diao-etal-2024-active}.

Newer strategies \textbf{leverage optimisation} to select a single representative set of samples. One possibility is to train a sample retriever, such as a separate scoring model~\citep{rubin-etal-2022-learning, luo2023dr, li-etal-2023-unified, wang2023learning, aimen2023leveraging} or define a surrogate scorer for modelling the sample's goodness and use multi-armed bandit to select the best exemplars~\citep{purohit-etal-2024-explora, purohit2025sample}. Another option is to leverage reinforcement learning~\citep{zhang-etal-2022-active, shum-etal-2023-automatic, scarlatos2023reticl}. Finally, the Datamodels approach trains a linear model to predict the performance gain of a set of samples to select the subset that would lead to the highest possible performance increase~\citep{ilyas2022datamodels, chang-jia-2023-data, jundi-lapesa-2022-translate, vilar-etal-2023-prompting}.

In this work, we investigate whether the combination of objectives from well-established single-property strategies can rival the selection from the strategies tailored specifically to in-context learning. Similar to the close works of~\citet{li-qiu-2023-finding, purohit-etal-2024-explora, purohit2025sample}, we select a single set of samples, but do not focus solely on in-context learning. In essence, we complement work by \citet{agarwal2021sensitivity} by exploring and effectively combining sample properties important for few-shot learning performance. Finally, our proposed method is inspired by the Datamodels approach~\citep{ilyas2022datamodels}, but performs the selection at the level of strategies instead of samples.
\section{ACSESS: Automatic Combination of Selection Strategies}
\label{sec:ACSESS}
Despite the large number of existing sample selection strategies, most remain constrained to a single objective, typically considering only one sample property. This narrow focus often leads to suboptimal selection, whereas greater performance gains can be achieved by appropriately combining multiple objectives that capture complementary sample properties. For example, the most informative sample that is hard to learn or is often forgotten may not contribute as much for few-shot methods that utilise gradient-based training. On the other hand, for in-context learning, the samples on the decision boundary, which are often hard to learn, may provide more benefit. 

Instead of focusing on a single property, we propose to select examples that are characterised by a balance of complementary properties, even when they may not represent the best sample of any given individual property. To accomplish this, we propose \textbf{ACSESS}, a method for \textbf{A}utomatic \textbf{C}ombination of \textbf{S}ampl\textbf{E} \textbf{S}election \textbf{S}trategies taking their complementarity into account. The method is composed of three stages: 1) \textit{definition of single-property strategies} that will be considered for the combination; 2) \textit{identification of a subset of well-performing and relevant strategies} that can improve the performance of few-shot learning; and then 3) \textit{combining the identified strategies} to identify the final set of the most beneficial samples.

To combine selection objectives, the ACSESS method assigns a \textit{selection score} for each sample, representing its expected contribution towards a higher performance. It is calculated as a weighted sum of scores provided by individual objectives, formally $score(x)=\sum_{s \in S} w_s * objective_{s}(x)$, where $\mathbf{S}$ represents the set of considered strategies, $w_s$ is the weight assigned to the strategy $s$, $x$ is the input sample to be scored, and $objective_s(x)$is the score assigned to the sample $x$ by strategy $s$ normalised to a $<0,1>$ interval. After assigning a score to each sample $x$, $N$ samples with the highest score are selected. The steps of the method are summarised in Algorithm~\ref{alg:method}, with further details, including discussion of the computation complexity and its optimisation, in Appendix~\ref{app:acsess-details}.

\begin{algorithm}[!b]
\caption{Automatic Combination of Sample Selection Objectives} \label{alg:method}
\begin{algorithmic}[1]
\small
\Require ${x \in X}$ - a set of available labelled samples
\Require $N$ - number of samples to choose

\State Define $\mathbf{S}$, a set of considered sample selection strategies
\State Apply forward selection, $\mathbf{S_{F}} = forward(N, \mathbf{S})$ (Alg. \ref{alg:forward})
\State Apply backward selection, $\mathbf{S_{B}} = backward(N, \mathbf{S})$ (Alg. \ref{alg:backward})
\State Calculate  performance of the few-shot model on the baseline setting, $perf$
\State Apply datamodels selection, $\mathbf{S_{D}}, weights_{D} = datamodels(N, \mathbf{S}, perf)$ (Alg. \ref{alg:datamodels})

\State Select final set of strategies $\mathbf{S_{final}} = \mathbf{S_{F}} \cap \mathbf{S_{B}} \cap \mathbf{S_{D}}$
\State ${w} = weights_{D}$ if weighted combination else ${1/|\mathbf{S_{final}}|}$
\State Calculate final set of scores for each sample $score(x) = \sum_{s \in S} w_s * objective_s(x)$
\State Select top $N$ samples based on $score(x)$
\end{algorithmic}
\end{algorithm}

\paragraph{Single-Property Strategies}

We focus on well-established strategies that select based on following properties: 1) \textit{informativeness}, or how informative the samples are for the model; 2) \textit{representativeness}, or how well the subset of samples represents the full dataset; and 3) \textit{learnability}, or how easy it is to obtain or retain the information contained in the samples. In this work, we specifically focus on strategies that work by assigning a score to each sample, sorting based on this score and selecting the samples with the best score. For \textit{informativeness}, this includes selecting samples based in their \textit{Similarity} and \textit{Diversity}, active learning strategies, such as \textit{Entropy}, \textit{Margin}, \textit{Least Confidence} and \textit{Loss}~\citep{park2022active}, and core-set selection including \textit{Contrastive Active Learning (CAL)}~\citep{margatina-etal-2021-active}, \textit{DeepFool}~\citep{ducoffe2018adversarial}, \textit{GraNd}~\citep{paul2021grand} and \textit{Graph-Cut}~\citep{iyer2013submodular, iyer2021submodular}. For representativeness, we consider \textit{Herding}~\citep{welling2009herding, chen2010herding}, \textit{KCenter}~\citep{sener2018active, agarwal2020contextual}, \textit{CRAIG}~\citep{mirzasoleiman2020coresets} and \textit{Glister}~\citep{killamsetty2021glister}. Finally, for learnability, we consider \textit{Forgetting}~\citep{toneva2018empirical} and \textit{Cartography}~\citep{swayamdipta-etal-2020-dataset, zhang-plank-2021-cartography-active}, selecting all of each, ambiguous, hard and a combination of easy and ambiguous samples (due to no obvious consensus on which are the best). More details regarding the strategies and their objectives are included in Appendix~\ref{app:sss-details}.

\paragraph{Identifying the Subset of Relevant Strategies}

In the second phase, the goal is to identify relevant strategies that, when combined, can select high-quality samples based on their complementary properties. Even though the optimal combination could be achieved by searching through all the combinations of strategies, such a search is infeasible as it requires extensive computation. Instead, we draw inspiration from methods commonly used for feature selection in traditional machine learning -- namely the \textit{forward} and \textit{backward} selection -- and for the first time apply them to the problem of sample selection. In addition, we adapt \textit{Datamodels}, introduced by~\citet{ilyas2022datamodels} for individual sample selection, to work on the level of strategies instead. Such an approach allows us to explore a more diverse set of possible strategy combinations as compared to forward/backward selection. All three of the methods are run independently of each other, each producing a separate set of strategies, i.e., $\mathbf{S_{F}}$, $\mathbf{S_{B}}$ and $\mathbf{S_{D}}$ respectively for forward, backward and datamodels selection. The final set of strategies used by our method is constructed as an intersection between these sets, i.e., $\mathbf{S_{final}} = \mathbf{S_{F}} \cap \mathbf{S_{B}} \cap \mathbf{S_{D}}$. This allows the method to identify the most impactful strategies (i.e., strategies identified by all of the methods), minimise the potential drawbacks of the individual methods (i.e., exploring also diverse combinations), and guarantee efficiency of the method, by identifying the lowest number of strategies possible.

\textbf{Forward Selection} iteratively adds strategies that result in the highest performance increase (similarly to forward feature selection). Starting with an empty subset, the method alternates between two steps. First, for each strategy not included in the subset of selected strategies, we determine the increase in performance the strategy would yield if added to the subset. We determine the increase by training the model with $N$ samples with the highest selection score determined by so far selected strategies (with weights uniformly distributed for all strategies, $w_s=1/|\mathbf{S_{temp}}|$). Second, the strategy that provides the highest positive increase in performance is added to the subset and the first step is repeated. If the addition of any strategy does not lead to a performance increase, the process ends, resulting in the subset of relevant selection strategies ($\mathbf{S_{F}}$) (see Algorithm~\ref{alg:forward} in Appendix~\ref{app:acsess-details} for more details). In essence, the forward selection iteratively adds the best-performing strategies to the set while the performance still increases.

\textbf{Backward Selection} iteratively removes strategies until the performance no longer increases (similarly to backward feature selection). The method works the same way as the forward selection, but starts with all selection strategies and removes them from the subset ($\mathbf{S_{B}}$) based on the performance increase such removal leads to (see Algorithm~\ref{alg:backward} in Appendix~\ref{app:acsess-details} for more details). In essence, the backward selection iteratively removes strategies that lead to lower performance until the performance no longer increases.

\textbf{Datamodels Selection} draws inspiration from Datamodels~\citep{ilyas2022datamodels, chang-jia-2023-data}. First, a set of $M$ (in our work $M=150$) random combinations of strategies is created and evaluated (using uniform combination). The random search is constrained to include each strategy at least 5 times to ensure sufficient coverage of each strategy. Afterwards, these combinations and their performance are used to create a simple regression dataset. The difference in performance to the baseline selection (the classic few-shot learning selection) is used as the target. As features, we use a multi-label presence vector of strategies, i.e., vector containing only zeros and ones, where 1 is assigned to elements corresponding to strategy that was used in the combination, and 0 for all other ones. The dataset is then used to train a regularised linear regression, LASSO~\citep{tibshirani1996regression}, which zeros out the weights for as many strategies as possible. The selection strategies with positive weights in the model represent the subset of relevant strategies ($\mathbf{S_D}$) (see Algorithm~\ref{alg:datamodels} in Appendix~\ref{app:acsess-details} for more details).

\paragraph{Combining the Identified Strategies}
After identifying the relevant selection strategies ($\mathbf{S_{final}}$), a weighted combination of objectives is used to perform the selection. The final set consists of $N$ samples with the highest selection score.

In this paper, we evaluate three separate weighting schemes. First, \textbf{uniform} combination, assigns the same weight to each of the identified strategies ($w_s={1/|\mathbf{S_{final}}|}$). This represents the simplified version of the ACSESS method that potentially leads to lower performance impact, but also lower computation costs and better transferability, as no further optimisation is required. This weighting scheme is also used for the previous step for the forward, backward and datamodels selection. Second, \textbf{weighted} combination, assigns a different weight to each strategy objective, prioritising specific sample properties beneficial for the specific few-shot approach. As the strategy weights for this weighted combination, we use the final trained weights of the \textit{Datamodels} linear model. As this weighting scheme is dataset and model-specific, it often leads to the highest performance impact, at the cost of requiring more extensive computation. For this reason, in this paper, we evaluate the strategy on a more aggregated level in terms of datasets -- see Appendix~\ref{app:acsess-details} for details. Finally, the \textbf{weighted combination with random selection} extends the \textit{weighted} combination by introducing a random element to it, as the random selection was identified as a strong baseline~\citep{park2022active, cegin-etal-2025-use}. For this weighting scheme, each sample is assigned an additional uniformly randomised score. To determine the weight of this score, the random selection is included as an additional strategy to the \textit{Datamodels} method.
\section{Experiments}
\label{sec:experiments}

\textbf{Datasets.} We focus on representative few-shot learning datasets composed of different tasks with different numbers of classes and with different characteristics, such as class imbalance or amount of noisy samples or labels. Specifically, we use 6 widely-adopted text datasets, composed of tasks with different number of classes and characteristics, specifically: \textbf{20 News Group}~\citep{LANG1995331} and \textbf{News Category}~\citep{misra2022news} for news category classification and \textbf{ATIS}~\citep{hemphill-etal-1990-atis}, \textbf{Facebook}~\citep{schuster-etal-2019-cross-lingual}, \textbf{HWU-64}~\citep{liu2021benchmarking} and \textbf{SNIPS}~\citep{coucke2018snips} for intent classification. For each dataset, we select a subset of up to 200 labelled samples per class to simulate the limited data setting if needed (the impact of smaller dataset sizes is analysed in Appendix~\ref{app:dataset_size_change_deag}). Furthermore, to evaluate the generalisability, we repeat the experiments on image data as well (see Appendix~\ref{app:image-results} for details).

\textbf{Few-shot Learning Approaches.} The evaluation is done in a \textbf{5-way 5-shot setting} using in-context learning on the text datasets and the gradient few-shot learning on both the text and image datasets. For in-context learning, we use the \textbf{Mistral-7B}~\citep{jiang2023mistral}, \textbf{Zephyr-7B}~\citep{tunstall2023zephyr}, \textbf{LLaMA-3.1-8B}~\citep{grattafiori2024llama}, \textbf{Gemma-3-4B}~\citep{gemma_2025} and \textbf{Qwen-2.5-7B}~\citep{qwen2.5} instruction-tuned large language models. For gradient few-shot learning, we use representative and widely-used few-shot learning approaches, specifically 1) meta-learning approaches \textbf{Prototypical Networks}~\citep{snell2017prototypical} and \textbf{MAML}~\citep{finn2017model}, and 2) \textbf{Few-Shot Fine-Tuning}~\citep{chen2019closer}.

\textbf{Baselines.} We compare all selected single-property selection strategies (see also  Table~\ref{tab:investigated-strategies} in Appendix~\ref{app:sss-details}) with 2 baselines: 1) \textbf{Classic selection}, where a new set of samples is randomly selected for each new task; and 2) \textbf{Random selection}, where only a single set of 5 samples per class is randomly selected once and used for every single task, representing a more realistic setting with limited budget. In-context learning is also compared with the recently proposed state-of-the-art in-context learning specific strategies \textbf{LENS}~\citep{li-qiu-2023-finding}, \textbf{Active Prompt}~\citep{diao-etal-2024-active}, \textbf{EXPLORA}~\citep{purohit-etal-2024-explora} and \textbf{CASE}~\citep{purohit2025sample}. 

Each selection strategy is run 10 times with different random seeds to determine their sensitivity to the effects of randomness (e.g., random initialisation or sample order). In addition, each experiment is repeated 5 times for different data splits (and the choice of the 200 labelled samples) and an average accuracy over these runs is reported (if not specified otherwise). Furthermore, each evaluation run is done on 600 randomly sampled tasks for gradient few-shot learning and 300 tasks for in-context learning (due to high computation costs of LLMs), where each task represents a random subset of 5 classes from the dataset (i.e., to construct the 5-way classification setting, following the standard few-shot learning methodology). As such, repeating the evaluation across multiple tasks and seeds deals with any possibility of overfitting and minimises the possible biases and other confounding factor in the evaluation. Further experiment setup details are in Appendix~\ref{app:experimental-setup-details}.

\subsection{Impact of Selection Strategies}
\begin{figure*}[t!]
    \centering
    \includegraphics[width=\linewidth]{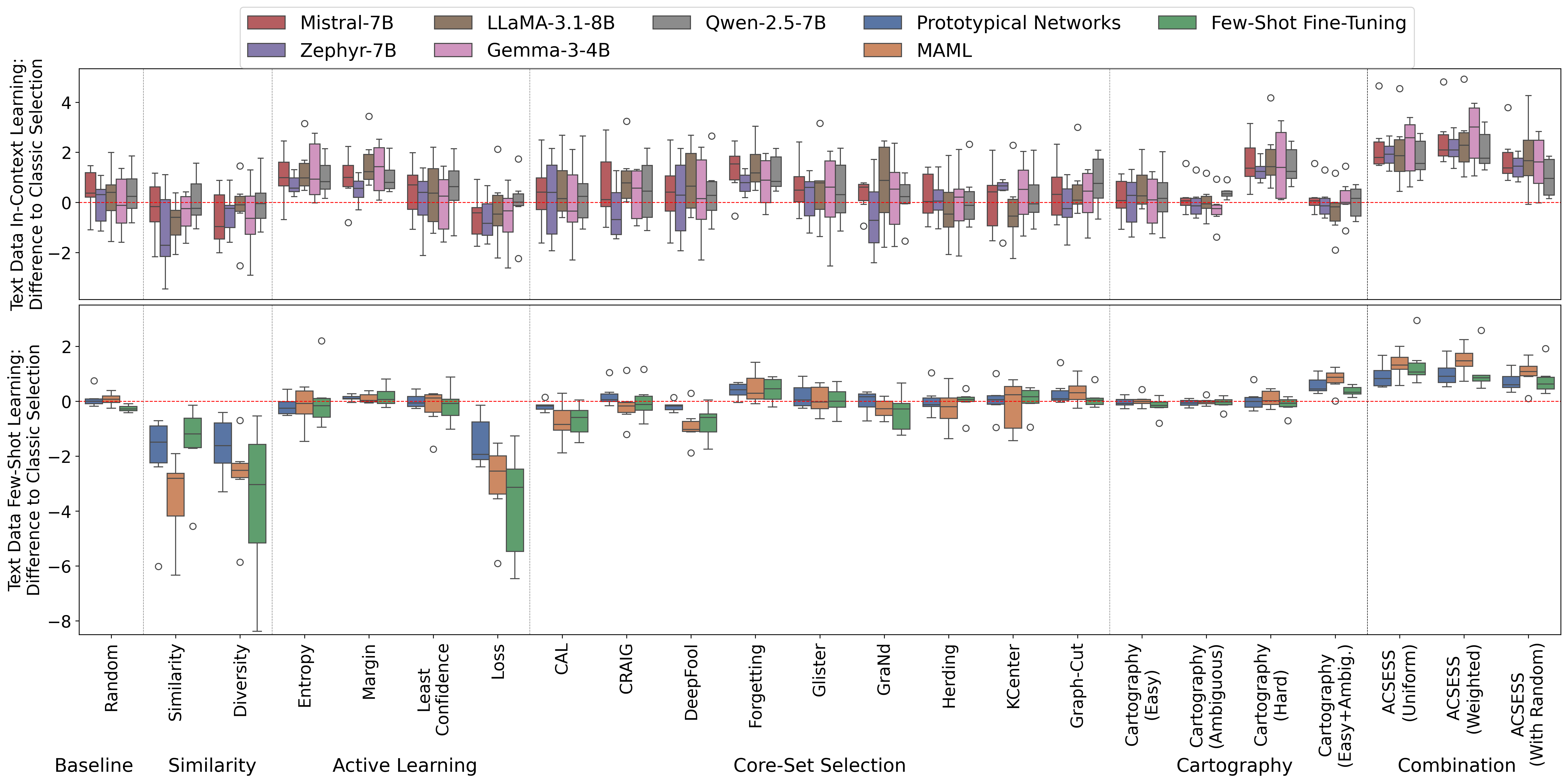}
    \caption{Benefit of the sample selection strategies, calculated as the difference in accuracy between the specific strategy and the classic few-shot selection (represented as the red dashed line), aggregated over the text datasets.}
    \label{fig:performance-increase}
\end{figure*}

In this section, our goal is to answer the following research question: \emph{\textbf{RQ1:} What is the impact of single-property selection strategies on the few-shot learning?} The objective is to analyse the performance increase of the different selection strategies across various datasets and few-shot learning approaches. The results are shown in Figure~\ref{fig:performance-increase} and Table~\ref{tab:strategies-diff-text} in Appendix~\ref{app:table-text}, aggregated over all the datasets (the results for image data are included in Appendix~\ref{app:image-results} and for individual datasets in supplementary material). We present the results separately for in-context learning and gradient few-shot learning, as we observed a strong dependence of the strategy benefit on the few-shot learning approach used. Additional ablation studies for the selection strategies are included in Appendix~\ref{app:additional_insights_dependence_sensitivity}.

\textbf{The majority of selection strategies fail to outperform the \textit{Classic selection} in case of gradient few-shot learning}. In most cases, the strategies lead to performance on par or even lower than the \textit{Classic selection} (for example diverse samples on text datasets). Only specific strategies consistently lead to better performance (e.g., combination of easy to learn and ambiguous samples, or \textit{Forgetting} on text datasets). As such, the \textbf{random sample selection represents a strong baseline} for the gradient few-shot learning.

\textbf{Sample selection is more beneficial for in-context learning.} Many of the selection strategies lead to increase in performance for in-context learning, while only few strategies, such as the popular selection based on \textit{similarity} or \textit{diversity}, under-perform the \textit{Classic selection}. At the same time, \textbf{the increase in performance is higher than with gradient few-shot learning}, although with a larger variance in results -- average increase of $1.5$ for in-context learning as compared to $0.5$ for gradient few-shot learning. In contrast to gradient few-shot learning, the hard to learn samples provide the most benefit for in-context learning. The reason may be that hard to learn samples provide a lot of information beneficial for the classification that cannot be easily leveraged in the gradient few-shot learning due to the information from the samples being hard to learn or retain. However, as there is no explicit "learning" in the in-context learning use of large language models, such samples can be better leveraged. As such, \textbf{learnability property is a strong indicator of quality for all few-shot learning approaches}. Selecting samples based on how easy (or hard for in-context learning) it is to learn them or how often they are forgotten consistently leads to the overall highest increase in performance.

\subsection{ACSESS Results and Characteristics}

\begin{figure}[tbh]
    \centering
    \includegraphics[width=\linewidth]{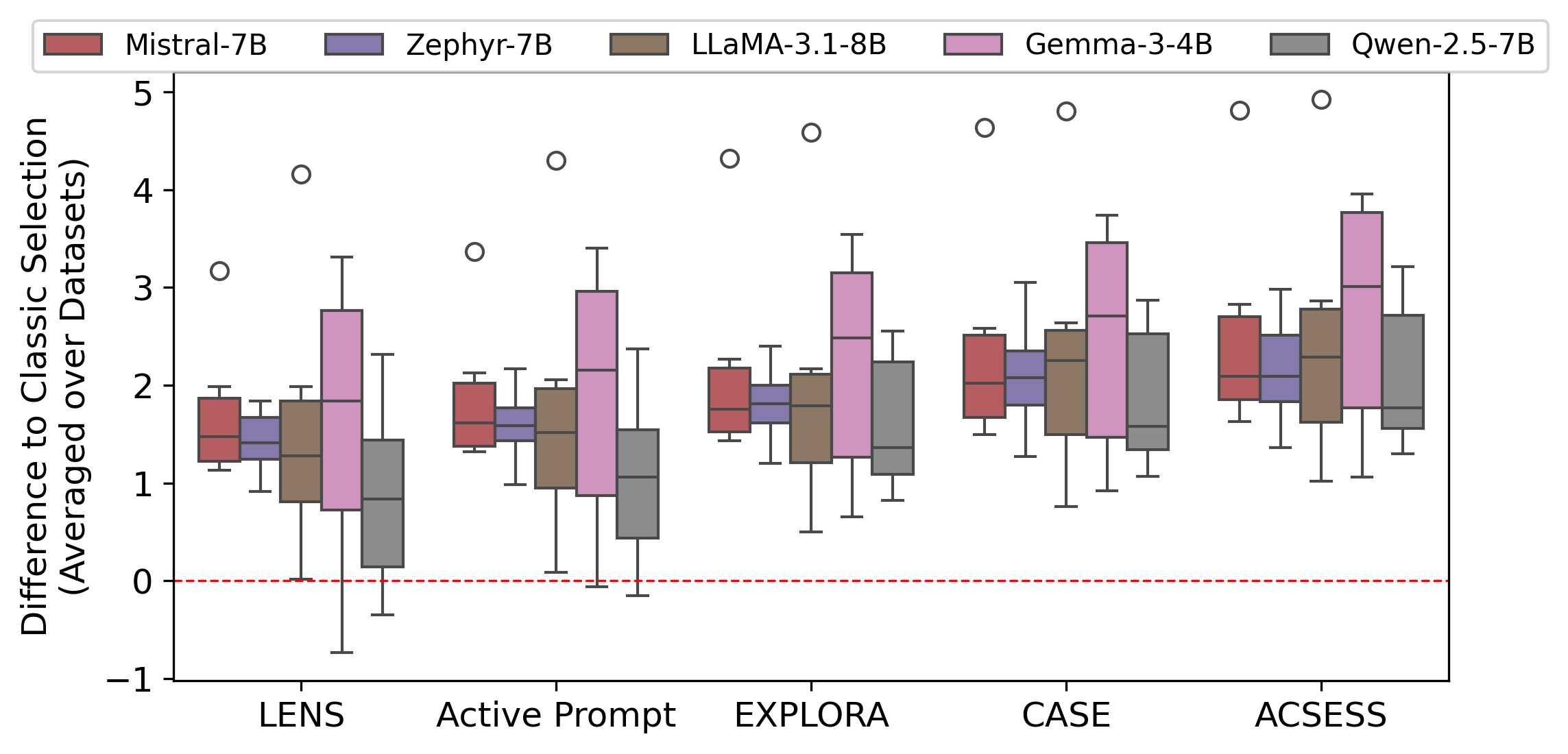}
    \caption{Comparison of the ACSESS method with in-context learning specific baselines.}
    \label{fig:combination-lens-diff}
\end{figure}

In this section, we answer the following research question: \emph{\textbf{RQ2:} How does the ACSESS method perform in comparison to the in-context learning specific baselines and single-property strategies?} Besides analysing and comparing the benefit with other strategies, we analyse the different weighting schemes of our method.

\textbf{The ACSESS method consistently leads to performance increases across all few-shot learning approaches.} Our proposed method outperforms all the investigated single-property strategies, leading to an average performance increase of up to $2.5$ percentage points for in-context learning and $1.8$ percentage points for gradient few-shot learning. Similar behaviour can be observed for the in-context specific baselines as well (LENS, Active Prompt, EXPLORA, CASE; see Table~\ref{tab:strategies-diff-text} and Figure~\ref{fig:combination-lens-diff} for comparison). Overall, the ACSESS method is able to outperform these baselines as well, achieving on average performance increase of $0.95$ percentage points over the worst performing LENS method and $0.16$ percentage points over the best performing CASE method. In all cases, the increase is statistically significant using the Wilcoxon signed-rank test. These results confirm that \textbf{combining simple and well-established selection strategies leads to performance on par or better than strategies specifically tailored for the task of sample selection for in-context learning.} Although the in-context learning specific strategies may show their benefit in specific setting (e.g., similarity based selection for translation), they do not generalise as well to other tasks.

\textbf{The ACSESS method identifies samples with complementary properties, with higher priority on learnability.} For the different types of few-shot learning (in-context vs. gradient learning), our method identifies and combines different set of selection strategies. 
In case of in-context learning, the combination of hard to learn samples (\textit{Cartography}) that are least often forgotten (\textit{Forgetting}) on the decision boundary (\textit{Margin}) and with highest \textit{Entropy} provide the most benefit. For the gradient few-shot learning approaches on text datasets, the combination of easy and ambiguous samples (\textit{Cartography}) that are least often forgotten (\textit{Forgetting}) on the decision boundary (\textit{Margin}) that provide additional informativeness from the \textit{Graph-Cut} strategy provides the most benefit. Further analysing the identified strategies and their weights, we find the \textbf{ACSESS method gives higher importance to samples that are \textit{learnable}, lower importance to \textit{informative} samples} (but with multiple sources for both), \textbf{while the strategies that focus on \textit{representativeness} are not included at all}. This further reinforces the finding that learnability is a strong indicator of quality for few-shot learning.

\textbf{The dataset and approach specific weighting leads to the best performance}. The only exception is the Few-Shot Fine-Tuning approach, where the uniform weighting shows the most benefit. At the same time, the weighting strategy that incorporates the random selection often leads to worse performance and higher variance in the results. Even though the uniform combination is computationally less expensive than the weighted combination, its performance increase is only slightly lower (average difference of $0.10-0.25$). In case of in-context learning, the uniform weighting is often on par with the best-performing CASE baslines. As such, the \textbf{uniform weighting represents a good trade-off} between the performance increase and the computation costs, making it in general a better choice.

\begin{figure*}[tbh]
    \centering
    \includegraphics[width=.935\linewidth]{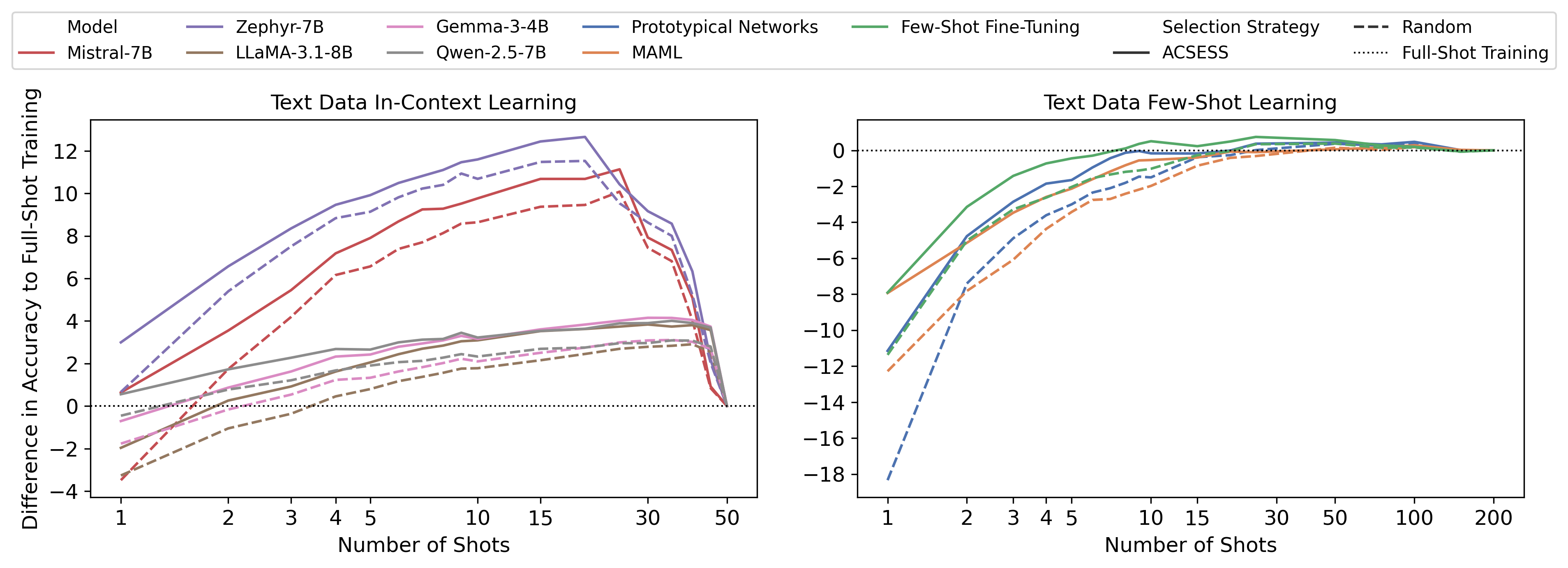}
    \caption{The change in benefit of selection strategies based on the number of shots used, aggregated over datasets. The change is calculated as a difference in accuracy to a setting using all available samples for training. When using a lower number of shots, the benefit of selection strategies over random selection is more significant. When using a larger number of shots, the benefit becomes negligible. In addition, the performance of full-shot training is matched, or in many cases exceeded, using smaller number of shots.}
    \label{fig:change-by-shots}
\end{figure*}

Based on these observations, we can draw practical recommendations. If the higher computation cost is not a problem, or one is interested in the highest possible increase in performance, the full ACSESS method should be run -- which includes running the full strategy identification for each dataset and model, and using the weighted combination of the strategies. The cost can be reduced by aggregating the identification over some of the dimensions, such as across multiple dataset or models, which leads to lower benefit. However, if extensive resource are not available, or we want to determine the potential benefit of sample selection, using a uniform combination of the \textit{Cartography} and \textit{Margin} (and in some cases \textit{Forgetting}) selection strategies provides a good approximation. This configuration does not incur any additional computation costs, while providing consistent improvement on all the datasets and models, which is only slightly lower than when using the model/dataset specific configuration.

\subsection{Effects of Variable Number of Shots and Dataset Size}

In this section, we answer the following research question: \emph{\textbf{RQ3:} How do the number of selected shots and the size of the dataset affect the benefit of sample selection?} We perform two ablation studies to investigate how the benefits of sample selection strategies change as we: 1) increase or decrease the number of selected shots; and 2) decrease the number of available labelled samples (i.e., the size of the dataset). For the first ablation study, we apply the \textit{ACSESS} method and \textit{Random} selection to select a different number of shots and compare their benefit with training on all available samples (results aggregated over all datasets are in Figure~\ref{fig:change-by-shots} and for individual datasets in Appendix~\ref{app:number_of_shots_deag}). For the second ablation study, we apply the \textit{ACSESS} method and \textit{Random} selection to select 5 samples per class from dataset subsets of different sizes and compare their benefit to the selection from the full dataset (results aggregated over all datasets are in Figure~\ref{fig:dataset_size_change} and for individual datasets in Appendix~\ref{app:dataset_size_change_deag}).

\begin{figure*}[tbh]
    \centering
    \includegraphics[width=.935\linewidth]{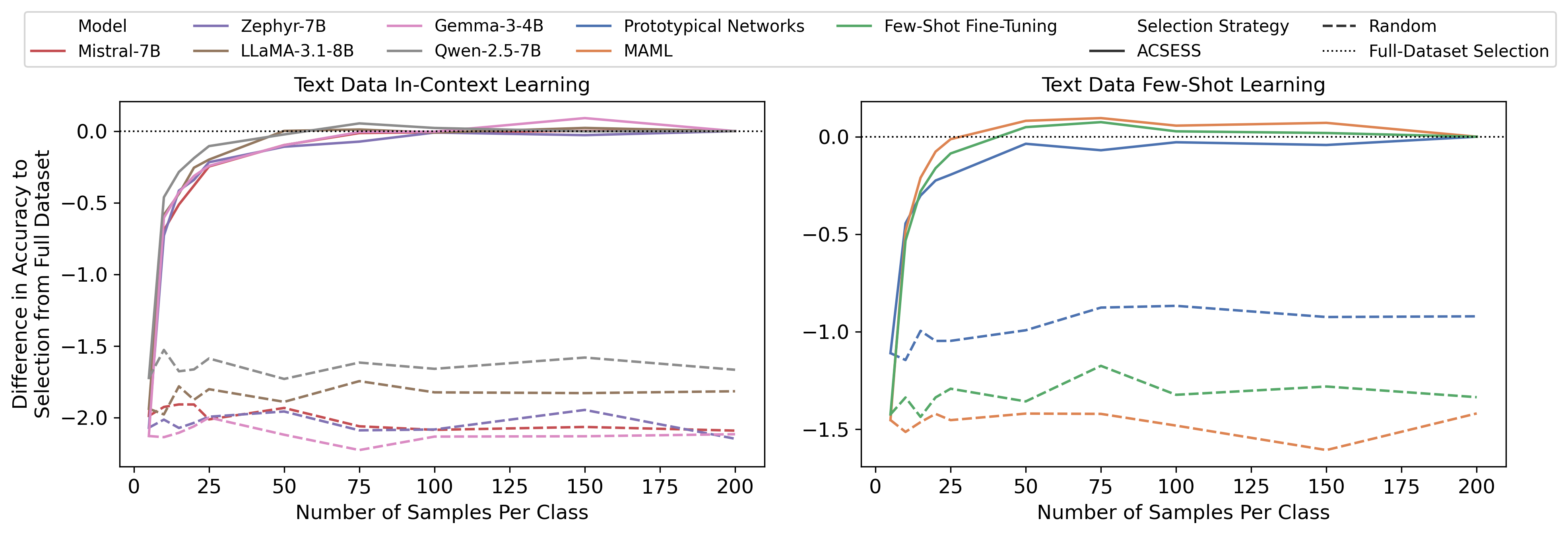}
    \caption{The difference in accuracy of selection strategies when using dataset subsets of different sizes, aggregated over the datasets. The difference is calculated as a comparison of the 5-shot selection using the dataset subset and the full dataset. Using only a fraction of dataset, the sample selection outperforms random baseline and quickly achieves similar performance to full dataset selection.}
    \label{fig:dataset_size_change}
\end{figure*}

\textbf{The selection strategies provide more benefit on lower number of shots, while their impact with further increase in number of shots is negligible after a certain point.} On the 1-shot setting, the difference in accuracy between the \textit{Random} selection and the \textit{ACSESS} strategy can be as high as $4$ or $7$ percentage points respectively for the in-context learning and gradient-based few-shot learning. In case of in-context learning, the impact is affected by the model used. For the older models (Mistral and Zephyr) the highest performance increase, as high as $10-12$ percentage points, is achieved at 20-25 shots. Afterwards, we observe a significant drop in performance, achieving similar performance to the 1-shot setting when using 50-shots. For newer models (LLaMA, Gemma and Qwen), we observe that increasing number of shots has only a negligible impact after around 5 shots, with the increase in performance around $2-3$ percentage points. However, we still observe a drop in performance when selecting 50 or more shots. The main reason for this drop-off may be the limited context size of the large language models, making the model degenerate to random guessing as the cross-over is text length dependent. For the gradient few-shot learning, the highest performance is achieved at lower number of shots ($20$) and then remains more or less constant (with slight decrease in some cases). In addition, increasing the number of selected shots in case of gradient-based few-shot learning leads to diminished impact, as the \textit{ACSESS} strategy regresses to random selection when selecting more than 30-40 shots. For in-context learning, we observe this behaviour only after the drop in performance. As such, the focus on \textbf{selecting high quality samples is paramount when selecting only a few labelled samples per class}, while the random selection can be safely used with higher number of shots. However, \textbf{labelling more samples and using them as additional shots may not always lead to better performance}.

\textbf{Sample selection is beneficial even when using dataset with low number of samples.} Using only 25\% (50 samples per class) for in-context learning or 10\% of the dataset (20 samples per class) for gradient few-shot learning, sample selection can still identify samples that lead to similar performance as is achieved when using the full dataset. Decreasing the size further, the benefit of sample selection starts to decrease as well. Using only 10 samples per class, the benefit decreases by approximately $20-40\%$ (e.g., from performance increase of $2$ percentage points to only $1.3$), and then quickly deteriorates to the \textit{Random} selection (as the number of samples to choose from is too low).
\section{Conclusion}

In this paper, we investigate whether the well-established sample selection objectives can rival the strategies tailored specifically to in-context learning. To this end, we propose the \textit{ACSESS} method for the automatic combination of the sample selection objectives to select samples with complementary properties and perform extensive evaluation and comparison of sample selection strategies on in-context learning and other few-shot learning approaches. The experimental results show that although the individual sample selection strategies may not always be beneficial, \textbf{their combination using our proposed method consistently leads to better performance, in many cases exceeding the performance from in-context learning specific strategies}. Curiously, we observe that the \textit{learnability} of the samples is more important for few-shot learning than their \textit{informativeness} or \textit{representativeness}, which is not often utilised in prior works. Based on these results, the uniform combination of \textit{Cartography}, \textit{Forgetting} and \textit{Margin} selection strategies is a good alternative transferable across all settings. Analysing the behaviour of the sample selection when changing the number of selected shots and the size of the dataset, we find that the sample selection provides the most benefit with a low number of shots, while with a higher number of shots, all strategies regress to random guessing. In addition, after a certain point, increasing the number of shots used for training no longer leads to improvement in the performance of few-shot learning, but may lead to a decrease in performance in some cases, especially for in-context learning. Overall, the benefit of selecting a subset of high-quality samples lies in improving the performance and reducing its variance for in-context learning, and in the reduction of computation cost, while providing a performance increase on the imbalanced and noisy datasets for the remaining few-shot learning approaches. Finally, even when the datasets contain only a few available samples per class, curating the samples further can often still lead to benefits, regardless of dataset characteristics or approach used.

\section*{Limitations}

\label{app:limitations}

To diminish undesirable randomness effects, we report the results as an average over 5 repeats of data splitting, 10 repeated runs of the selection strategies (resulting in 50 overall evaluation runs), with each run evaluated on 600 (or 300 for in-context learning) tasks. Such setting limits the variance in results, but may also obscure the benefit of selection strategies to a certain extent. On the other hand, although we focus only on a 5-way few-shot learning setting (using only 5 classes per task), we do not expect significant change in result for most of the few-shot learning approaches when using higher number of classes (such as 10-way or 20-way classification). The only exception is the in-context learning, where we observed the dependence on input size and context width of the models. With higher number of classes, the observed performance of large language models would start to deteriorate to the performance that can be achieved with a low number of shots. In this case, using larger model sizes may lead to better utilisation of the samples and higher performance increase from the selection strategies. 

Similarly to other research papers focusing on sample selection and few-shot learning via in-context learning, we assume an availability of large-enough annotated dataset from which we select samples using different strategies -- we use up to 200 samples per class for each dataset for the main experiments. As such, we also have larger validation or held-out set available for evaluating the selection strategies and hyperparameters for our proposed methods or for the approaches themselves. If we considered the true few-shot learning setting, where we have only unlabelled data available, only the active learning strategies (i.e., Entropy, Margin and Least Confidence) could be used, as they work with unlabelled data only, or a different approach for sample selection should be employed, such as using weak-labels to select the best samples, or assign every single class to all the labels and perform the sample selection that way to select both best performing samples as well as the class for them (as it is done by~\citet{chang-jia-2023-data}). \textbf{We address this (potentially most significant) limitation by running ablation studies to observe the benefit of sample selection} (using our proposed method and random selection) \textbf{when using smaller subsets of data} (i.e., selecting samples using only fraction of the dataset) \textbf{or using different number of shots per class.}

Before the experiments, we perform basic prompt-engineering for the in-context learning approaches and run hyperparameter optimisation for the gradient few-shot learning models. We follow the recommendations for good prompt engineering, using the meta-tags defined for each model (which lead to better performance of the models), while also not using overly-simple or overly-complicated prompts. The prompts we use are created based on dataset description, the prompts used in related works, our own experience, and the formats recommended for different models (e.g., taking inspiration from~\citet{sun2023pushing}). As we do not perform any extensive or automatic prompt-engineering, these prompts may lead to lowered performance and in some cases also lower impact of the sample selection for the few-shot learning setting. However, running an extensive prompt-engineering would introduce another controllable parameter which could potentially introduce further bias into the results (e.g., different selection strategies showing different impact for different prompt formats).

\section*{Ethical Considerations}
\label{app:ethical-cosiderations}

The experiments in this paper work with publicly available datasets, citing the original authors of each dataset (with the exception of the datasets contained in MetaAlbum, where we cite the creators of the aggregated dataset). All the datasets are used in accordance to their license and terms and conditions of use. We do not work with any personally identifiable information or offensive content and perform no crowdsourcing nor further data annotation. In addition, we are not aware of any potential ethical harms or negative societal impacts of our work, apart from the ones related to the advancement of the field of Machine Learning and Few-Shot learning (including large language models and in-context learning), such as the use of large amount of computation resources. Finally, we follow the license terms for all the models we use -- all models and datasets allow their use as part of research. The large language models (Mistral, Zephyr, LLaMA, Gemma and Qwen), which we use for the experiments, may contain biases and may generate potentially offensive or harmful content. However, the authors of the models already reduce this potential bias as much as possible when training the models, while at the same time we limit the output to few tokens and do not release any output of these models, which should further reduce the potential bias and negative impact.

\paragraph{Impact Statement: CO2 Emissions Related to Experiments} 
The experiments presented in this paper used larger amount of compute resources in order to evaluate all the sample selection strategies for all the different few-shot learning approaches. Overall, all the experiments (including preliminary experiments whose results are not reported in this paper) were conducted using a private infrastructure, which has a carbon efficiency of 0.432 kgCO$_2$eq/kWh. A cumulative of approximately 2500 hours of computation was performed on hardware of type A100 PCIe 40GB (TDP of 250W). Total emissions are estimated to be 270 kgCO$_2$eq of which 0 percent were directly offset. These estimations were conducted using the \href{https://mlco2.github.io/impact#compute}{MachineLearning Impact calculator}. Whenever possible, we tried to reduce the compute resources used as much as possible. The most compute resources were used by the large language models -- Mistral-7B, Zephyr-7B, LLaMA-3.1-8B~\citep{grattafiori2024llama}, Gemma-3-4B~\citep{gemma_2025} and Qwen-2.5-7B, and their further tuning. To reduce the computation costs, we have opted for using their smaller version and in cases when the further tuning is required, we fine-tune only the last classification layer (keeping the rest frozen). In addition, for gradient few-shot learning approaches, we opt for smaller models that do not require such an extensive computation costs, while still providing relevant results. Finally, whenever possible, we reuse the already performed computations for the selection strategies as well as using the compute optimised version of our proposed method (more information included in Appendix~\ref{app:acsess-details}), such as generating the feature representation for the different samples only a single time and re-using it whenever possible (i.e., using pre-trained models without further tuning for these representations).

\section*{Acknowledgments}

This work was partially supported by the projects funded by the European Union under the EU Horizon 2020: \textit{TAILOR}, GA No. \href{https://doi.org/10.3030/952215}{952215}, by the European Union under the Horizon Europe: \textit{vera.ai}, GA No. \href{https://doi.org/10.3030/101070093}{101070093}, \textit{AI-CODE}, GA No. \href{https://cordis.europa.eu/project/id/101135437}{101135437}; and by the Slovak Research and Development Agency: \textit{Modermed}, GA No. APVV-22-0414.

The presented results were obtained using the computational resources procured in the national project \textit{National competence centre for high performance computing} (project code: 311070AKF2) funded by European Regional Development Fund, EU Structural Funds Informatization of Society, Operational Program Integrated Infrastructure.


\bibliography{references}

@inproceedings{li-qiu-2023-finding,
    title = "Finding Support Examples for In-Context Learning",
    author = "Li, Xiaonan  and
      Qiu, Xipeng",
    booktitle = "Findings of the Association for Computational Linguistics: EMNLP 2023",
    month = dec,
    year = "2023",
    address = "Singapore",
    publisher = "ACL",
    doi = "10.18653/v1/2023.findings-emnlp.411",
    pages = "6219--6235",
}

@inproceedings{meta-album-2022,
    title={Meta-Album: Multi-domain Meta-Dataset for Few-Shot Image Classification},
    author={Ullah, Ihsan and Carrion, Dustin and Escalera, Sergio and Guyon, Isabelle M and Huisman, Mike and Mohr, Felix and van Rijn, Jan N and Sun, Haozhe and Vanschoren, Joaquin and Vu, Phan Anh},
    booktitle={Thirty-sixth Conference on Neural Information Processing Systems Datasets and Benchmarks Track},
    year = {2022}
}

@incollection{LANG1995331,
title = {NewsWeeder: Learning to Filter Netnews},
booktitle = {Machine Learning Proceedings 1995},
publisher = {Morgan Kaufmann},
address = {San Francisco (CA)},
pages = {331-339},
year = {1995},
isbn = {978-1-55860-377-6},
doi = {https://doi.org/10.1016/B978-1-55860-377-6.50048-7},
author = {Ken Lang}
}

@article{misra2022news,
  title={News category dataset},
  author={Misra, Rishabh},
  journal={arXiv preprint arXiv:2209.11429},
  year={2022}
}

@inproceedings{hemphill-etal-1990-atis,
    title = "The {ATIS} Spoken Language Systems Pilot Corpus",
    author = "Hemphill, Charles T.  and
      Godfrey, John J.  and
      Doddington, George R.",
    booktitle = "Speech and Natural Language: Proceedings of a Workshop Held at Hidden Valley, {P}ennsylvania, June 24-27,1990",
    year = "1990",
}

@inproceedings{schuster-etal-2019-cross-lingual,
    title = "Cross-lingual Transfer Learning for Multilingual Task Oriented Dialog",
    author = "Schuster, Sebastian  and
      Gupta, Sonal  and
      Shah, Rushin  and
      Lewis, Mike",
    booktitle = "Proceedings of the 2019 Conference of the North {A}merican Chapter of the Association for Computational Linguistics: Human Language Technologies, Volume 1 (Long and Short Papers)",
    month = jun,
    year = "2019",
    address = "Minneapolis, Minnesota",
    publisher = "Association for Computational Linguistics",
    doi = "10.18653/v1/N19-1380",
    pages = "3795--3805",
}

@inproceedings{liu2021benchmarking,
  title={Benchmarking natural language understanding services for building conversational agents},
  author={Liu, Xingkun and Eshghi, Arash and Swietojanski, Pawel and Rieser, Verena},
  booktitle={Increasing Naturalness and Flexibility in Spoken Dialogue Interaction: 10th International Workshop on Spoken Dialogue Systems},
  pages={165--183},
  year={2021},
  organization={Springer}
}

@article{coucke2018snips,
  title={Snips voice platform: an embedded spoken language understanding system for private-by-design voice interfaces},
  author={Coucke, Alice and Saade, Alaa and Ball, Adrien and Bluche, Th{\'e}odore and Caulier, Alexandre and Leroy, David and Doumouro, Cl{\'e}ment and Gisselbrecht, Thibault and Caltagirone, Francesco and Lavril, Thibaut and others},
  journal={arXiv preprint arXiv:1805.10190},
  year={2018}
}

@article{snell2017prototypical,
  title={Prototypical networks for few-shot learning},
  author={Snell, Jake and Swersky, Kevin and Zemel, Richard},
  journal={Advances in neural information processing systems},
  volume={30},
  year={2017}
}

@inproceedings{chen2019closer,
  title={A Closer Look at Few-shot Classification},
  author={Chen, Wei-Yu and Liu, Yen-Cheng and Kira, Zsolt and Wang, Yu-Chiang Frank and Huang, Jia-Bin},
  booktitle={International Conference on Learning Representations},
  year={2019}
}

@inproceedings{finn2017model,
  title={Model-agnostic meta-learning for fast adaptation of deep networks},
  author={Finn, Chelsea and Abbeel, Pieter and Levine, Sergey},
  booktitle={International conference on machine learning},
  pages={1126--1135},
  year={2017},
  organization={PMLR}
}

@article{jiang2023mistral,
  title={Mistral 7B},
  author={Jiang, Albert Q and Sablayrolles, Alexandre and Mensch, Arthur and Bamford, Chris and Chaplot, Devendra Singh and Casas, Diego de las and Bressand, Florian and Lengyel, Gianna and Lample, Guillaume and Saulnier, Lucile and others},
  journal={arXiv preprint arXiv:2310.06825},
  year={2023}
}

@article{tunstall2023zephyr,
  title={Zephyr: Direct distillation of lm alignment},
  author={Tunstall, Lewis and Beeching, Edward and Lambert, Nathan and Rajani, Nazneen and Rasul, Kashif and Belkada, Younes and Huang, Shengyi and von Werra, Leandro and Fourrier, Cl{\'e}mentine and Habib, Nathan and others},
  journal={arXiv preprint arXiv:2310.16944},
  year={2023}
}

@inproceedings{coleman2020Selection,
title={Selection via Proxy: Efficient Data Selection for Deep Learning},
author={Cody Coleman and Christopher Yeh and Stephen Mussmann and Baharan Mirzasoleiman and Peter Bailis and Percy Liang and Jure Leskovec and Matei Zaharia},
booktitle={International Conference on Learning Representations},
year={2020},
}

@inproceedings{park2022active,
title={Active Learning is a Strong Baseline for Data Subset Selection},
author={Dongmin Park and Dimitris Papailiopoulos and Kangwook Lee},
booktitle={Has it Trained Yet? NeurIPS 2022 Workshop},
year={2022},
}

@article{ren2021survey,
author = {Ren, Pengzhen and Xiao, Yun and Chang, Xiaojun and Huang, Po-Yao and Li, Zhihui and Gupta, Brij B. and Chen, Xiaojiang and Wang, Xin},
title = {A Survey of Deep Active Learning},
year = {2021},
issue_date = {December 2022},
publisher = {Association for Computing Machinery},
address = {New York, NY, USA},
volume = {54},
number = {9},
issn = {0360-0300},
doi = {10.1145/3472291},
journal = {ACM Comput. Surv.},
month = {oct},
articleno = {180},
numpages = {40}
}

@inproceedings{margatina-etal-2021-active,
    title = "Active Learning by Acquiring Contrastive Examples",
    author = {Margatina, Katerina  and
      Vernikos, Giorgos  and
      Barrault, Lo{\"\i}c  and
      Aletras, Nikolaos},
    booktitle = "Proceedings of the 2021 Conference on Empirical Methods in Natural Language Processing",
    month = nov,
    year = "2021",
    address = "Online and Punta Cana, Dominican Republic",
    publisher = "ACL",
    doi = "10.18653/v1/2021.emnlp-main.51",
    pages = "650--663",
}

@article{ducoffe2018adversarial,
  title={Adversarial active learning for deep networks: a margin based approach},
  author={Ducoffe, Melanie and Precioso, Frederic},
  journal={arXiv preprint arXiv:1802.09841},
  year={2018}
}

@inproceedings{iyer2013submodular,
 author = {Iyer, Rishabh K and Bilmes, Jeff A},
 booktitle = {Advances in Neural Information Processing Systems},
 pages = {},
 publisher = {Curran Associates, Inc.},
 title = {Submodular Optimization with Submodular Cover and Submodular Knapsack Constraints},
 volume = {26},
 year = {2013}
}

@InProceedings{iyer2021submodular,
  title = 	 {Submodular combinatorial information measures with applications in machine learning},
  author =       {Iyer, Rishabh and Khargoankar, Ninad and Bilmes, Jeff and Asanani, Himanshu},
  booktitle = 	 {Proceedings of the 32nd International Conference on Algorithmic Learning Theory},
  pages = 	 {722--754},
  year = 	 {2021},
  volume = 	 {132},
  series = 	 {Proceedings of Machine Learning Research},
  month = 	 {16--19 Mar},
  publisher =    {PMLR}
}

@inproceedings{paul2021grand,
 author = {Paul, Mansheej and Ganguli, Surya and Dziugaite, Gintare Karolina},
 booktitle = {Advances in Neural Information Processing Systems},
 pages = {20596--20607},
 publisher = {Curran Associates, Inc.},
 title = {Deep Learning on a Data Diet: Finding Important Examples Early in Training},
 volume = {34},
 year = {2021}
}

@inproceedings{welling2009herding,
author = {Welling, Max},
title = {Herding Dynamical Weights to Learn},
year = {2009},
isbn = {9781605585161},
publisher = {ACM},
address = {New York, NY, USA},
doi = {10.1145/1553374.1553517},
booktitle = {Proceedings of the 26th Annual International Conference on Machine Learning},
pages = {1121–1128},
numpages = {8},
location = {Montreal, Quebec, Canada},
series = {ICML '09}
}

@inproceedings{chen2010herding,
author = {Chen, Yutian and Welling, Max and Smola, Alex},
title = {Super-Samples from Kernel Herding},
year = {2010},
isbn = {9780974903965},
publisher = {AUAI Press},
address = {Arlington, Virginia, USA},
booktitle = {Proceedings of the Twenty-Sixth Conference on Uncertainty in Artificial Intelligence},
pages = {109–116},
numpages = {8},
location = {Catalina Island, CA},
series = {UAI'10}
}

@inproceedings{sener2018active,
  title={Active Learning for Convolutional Neural Networks: A Core-Set Approach},
  author={Sener, Ozan and Savarese, Silvio},
  booktitle={International Conference on Learning Representations},
  year={2018}
}

@inproceedings{agarwal2020contextual,
  title={Contextual diversity for active learning},
  author={Agarwal, Sharat and Arora, Himanshu and Anand, Saket and Arora, Chetan},
  booktitle={Computer Vision--ECCV 2020: 16th European Conference, Glasgow, UK, August 23--28, 2020, Proceedings, Part XVI 16},
  pages={137--153},
  year={2020},
  organization={Springer}
}

@inproceedings{mirzasoleiman2020coresets,
  title={Coresets for data-efficient training of machine learning models},
  author={Mirzasoleiman, Baharan and Bilmes, Jeff and Leskovec, Jure},
  booktitle={International Conference on Machine Learning},
  pages={6950--6960},
  year={2020},
  organization={PMLR}
}

@inproceedings{killamsetty2021glister,
  title={Glister: Generalization based data subset selection for efficient and robust learning},
  author={Killamsetty, Krishnateja and Sivasubramanian, Durga and Ramakrishnan, Ganesh and Iyer, Rishabh},
  booktitle={Proceedings of the AAAI Conference on Artificial Intelligence},
  volume={35},
  pages={8110--8118},
  year={2021}
}

@inproceedings{toneva2018empirical,
  title={An Empirical Study of Example Forgetting during Deep Neural Network Learning},
  author={Toneva, Mariya and Sordoni, Alessandro and des Combes, Remi Tachet and Trischler, Adam and Bengio, Yoshua and Gordon, Geoffrey J},
  booktitle={International Conference on Learning Representations},
  year={2018}
}

@inproceedings{swayamdipta-etal-2020-dataset,
    title = "Dataset Cartography: Mapping and Diagnosing Datasets with Training Dynamics",
    author = "Swayamdipta, Swabha  and
      Schwartz, Roy  and
      Lourie, Nicholas  and
      Wang, Yizhong  and
      Hajishirzi, Hannaneh  and
      Smith, Noah A.  and
      Choi, Yejin",
    booktitle = "Proceedings of the 2020 Conference on Empirical Methods in Natural Language Processing (EMNLP)",
    month = nov,
    year = "2020",
    address = "Online",
    publisher = "Association for Computational Linguistics",
    doi = "10.18653/v1/2020.emnlp-main.746",
    pages = "9275--9293",
}

@inproceedings{zhang-plank-2021-cartography-active,
    title = "Cartography Active Learning",
    author = "Zhang, Mike  and
      Plank, Barbara",
    booktitle = "Findings of the Association for Computational Linguistics: EMNLP 2021",
    month = nov,
    year = "2021",
    address = "Punta Cana, Dominican Republic",
    publisher = "ACL",
    doi = "10.18653/v1/2021.findings-emnlp.36",
    pages = "395--406",
}

@inproceedings{ilyas2022datamodels,
  title={Datamodels: Predicting Predictions from Training Data},
  author={Ilyas, Andrew and Park, Sung Min and Engstrom, Logan and Leclerc, Guillaume and Madry, Aleksander},
  booktitle={Proceedings of the 39th International Conference on Machine Learning},
  year={2022}
}

@inproceedings{chang-jia-2023-data,
    title = "Data Curation Alone Can Stabilize In-context Learning",
    author = "Chang, Ting-Yun  and
      Jia, Robin",
    booktitle = "Proceedings of the 61st Annual Meeting of the Association for Computational Linguistics (Volume 1: Long Papers)",
    month = jul,
    year = "2023",
    address = "Toronto, Canada",
    publisher = "Association for Computational Linguistics",
    doi = "10.18653/v1/2023.acl-long.452",
    pages = "8123--8144",
}

@inproceedings{vilar-etal-2023-prompting,
    title = "Prompting {P}a{LM} for Translation: Assessing Strategies and Performance",
    author = "Vilar, David  and
      Freitag, Markus  and
      Cherry, Colin  and
      Luo, Jiaming  and
      Ratnakar, Viresh  and
      Foster, George",
    booktitle = "Proceedings of the 61st Annual Meeting of the Association for Computational Linguistics (Volume 1: Long Papers)",
    month = jul,
    year = "2023",
    address = "Toronto, Canada",
    publisher = "Association for Computational Linguistics",
    doi = "10.18653/v1/2023.acl-long.859",
    pages = "15406--15427",
}

@article{yu2023dataset,
  title={Dataset distillation: A comprehensive review},
  author={Yu, Ruonan and Liu, Songhua and Wang, Xinchao},
  journal={arXiv preprint arXiv:2301.07014},
  year={2023}
}

@inproceedings{guo2022deepcore,
  title={Deepcore: A comprehensive library for coreset selection in deep learning},
  author={Guo, Chengcheng and Zhao, Bo and Bai, Yanbing},
  booktitle={International Conference on Database and Expert Systems Applications},
  pages={181--195},
  year={2022},
  organization={Springer}
}

@article{sun2023pushing,
  title={Pushing the Limits of ChatGPT on NLP Tasks},
  author={Sun, Xiaofei and Dong, Linfeng and Li, Xiaoya and Wan, Zhen and Wang, Shuhe and Zhang, Tianwei and Li, Jiwei and Cheng, Fei and Lyu, Lingjuan and Wu, Fei and others},
  journal={arXiv preprint arXiv:2306.09719},
  year={2023}
}

@inproceedings{zhang-etal-2022-active,
    title = "Active Example Selection for In-Context Learning",
    author = "Zhang, Yiming  and
      Feng, Shi  and
      Tan, Chenhao",
    booktitle = "Proceedings of the 2022 Conference on Empirical Methods in Natural Language Processing",
    month = dec,
    year = "2022",
    address = "Abu Dhabi, United Arab Emirates",
    publisher = "Association for Computational Linguistics",
    doi = "10.18653/v1/2022.emnlp-main.622",
    pages = "9134--9148",
}

@article{pecher2023effects,
author = {Pecher, Branislav and Srba, Ivan and Bielikova, Maria},
title = {A Survey on Stability of Learning with Limited Labelled Data and its Sensitivity to the Effects of Randomness},
year = {2024},
issue_date = {January 2025},
publisher = {Association for Computing Machinery},
address = {New York, NY, USA},
volume = {57},
number = {1},
issn = {0360-0300},
url = {https://doi.org/10.1145/3691339},
doi = {10.1145/3691339},
journal = {ACM Comput. Surv.},
month = oct,
articleno = {19},
numpages = {40}
}

@inproceedings{shum-etal-2023-automatic,
    title = "Automatic Prompt Augmentation and Selection with Chain-of-Thought from Labeled Data",
    author = "Shum, Kashun  and
      Diao, Shizhe  and
      Zhang, Tong",
    booktitle = "Findings of the Association for Computational Linguistics: EMNLP 2023",
    month = dec,
    year = "2023",
    address = "Singapore",
    publisher = "Association for Computational Linguistics",
    doi = "10.18653/v1/2023.findings-emnlp.811",
    pages = "12113--12139",
}

@article{ma2023fairness,
  title={Fairness-guided Few-shot Prompting for Large Language Models},
  author={Ma, Huan and Zhang, Changqing and Bian, Yatao and Liu, Lemao and Zhang, Zhirui and Zhao, Peilin and Zhang, Shu and Fu, Huazhu and Hu, Qinghua and Wu, Bingzhe},
  journal={arXiv preprint arXiv:2303.13217},
  year={2023}
}

@misc{wu2023selfadaptive,
      title={Self-Adaptive In-Context Learning: An Information Compression Perspective for In-Context Example Selection and Ordering}, 
      author={Zhiyong Wu and Yaoxiang Wang and Jiacheng Ye and Lingpeng Kong},
      year={2023},
      eprint={2212.10375},
      archivePrefix={arXiv},
      primaryClass={cs.CL}
}

@article{qin2023context,
  title={In-context learning with iterative demonstration selection},
  author={Qin, Chengwei and Zhang, Aston and Dagar, Anirudh and Ye, Wenming},
  journal={arXiv preprint arXiv:2310.09881},
  year={2023}
}

@inproceedings{liu2023towards,
  title={Towards Informative Few-Shot Prompt with Maximum Information Gain for In-Context Learning},
  author={Liu, Hongfu and Wang, Ye},
  booktitle={Findings of the Association for Computational Linguistics: EMNLP 2023},
  pages={15825--15838},
  year={2023}
}

@inproceedings{li-etal-2023-unified,
    title = "Unified Demonstration Retriever for In-Context Learning",
    author = "Li, Xiaonan  and
      Lv, Kai  and
      Yan, Hang  and
      Lin, Tianyang  and
      Zhu, Wei  and
      Ni, Yuan  and
      Xie, Guotong  and
      Wang, Xiaoling  and
      Qiu, Xipeng",
    booktitle = "Proceedings of the 61st Annual Meeting of the Association for Computational Linguistics (Volume 1: Long Papers)",
    month = jul,
    year = "2023",
    address = "Toronto, Canada",
    publisher = "Association for Computational Linguistics",
    doi = "10.18653/v1/2023.acl-long.256",
    pages = "4644--4668",
}

@article{mavromatis2023examples,
  title={Which Examples to Annotate for In-Context Learning? Towards Effective and Efficient Selection},
  author={Mavromatis, Costas and Srinivasan, Balasubramaniam and Shen, Zhengyuan and Zhang, Jiani and Rangwala, Huzefa and Faloutsos, Christos and Karypis, George},
  journal={arXiv preprint arXiv:2310.20046},
  year={2023}
}

@inproceedings{liu-etal-2022-makes,
    title = "What Makes Good In-Context Examples for {GPT}-3?",
    author = "Liu, Jiachang  and
      Shen, Dinghan  and
      Zhang, Yizhe  and
      Dolan, Bill  and
      Carin, Lawrence  and
      Chen, Weizhu",
    booktitle = "Proceedings of Deep Learning Inside Out (DeeLIO 2022): The 3rd Workshop on Knowledge Extraction and Integration for Deep Learning Architectures",
    month = may,
    year = "2022",
    address = "Dublin, Ireland and Online",
    publisher = "Association for Computational Linguistics",
    doi = "10.18653/v1/2022.deelio-1.10",
    pages = "100--114",
}

@inproceedings{an-etal-2023-skill,
    title = "Skill-Based Few-Shot Selection for In-Context Learning",
    author = "An, Shengnan  and
      Zhou, Bo  and
      Lin, Zeqi  and
      Fu, Qiang  and
      Chen, Bei  and
      Zheng, Nanning  and
      Chen, Weizhu  and
      Lou, Jian-Guang",
    booktitle = "Proceedings of the 2023 Conference on Empirical Methods in Natural Language Processing",
    month = dec,
    year = "2023",
    address = "Singapore",
    publisher = "ACL",
    doi = "10.18653/v1/2023.emnlp-main.831",
    pages = "13472--13492",
}

@inproceedings{zemlyanskiy-etal-2022-generate,
    title = "Generate-and-Retrieve: Use Your Predictions to Improve Retrieval for Semantic Parsing",
    author = "Zemlyanskiy, Yury  and
      de Jong, Michiel  and
      Ainslie, Joshua  and
      Pasupat, Panupong  and
      Shaw, Peter  and
      Qiu, Linlu  and
      Sanghai, Sumit  and
      Sha, Fei",
    booktitle = "Proceedings of the 29th International Conference on Computational Linguistics",
    month = oct,
    year = "2022",
    address = "Gyeongju, Republic of Korea",
    publisher = "International Committee on Computational Linguistics",
    pages = "4946--4951",
}

@inproceedings{gupta-etal-2022-retronlu,
    title = "{R}etro{NLU}: Retrieval Augmented Task-Oriented Semantic Parsing",
    author = "Gupta, Vivek  and
      Shrivastava, Akshat  and
      Sagar, Adithya  and
      Aghajanyan, Armen  and
      Savenkov, Denis",
    booktitle = "Proceedings of the 4th Workshop on NLP for Conversational AI",
    month = may,
    year = "2022",
    address = "Dublin, Ireland",
    publisher = "Association for Computational Linguistics",
    doi = "10.18653/v1/2022.nlp4convai-1.15",
    pages = "184--196",
}

@inproceedings{pasupat-etal-2021-controllable,
    title = "Controllable Semantic Parsing via Retrieval Augmentation",
    author = "Pasupat, Panupong  and
      Zhang, Yuan  and
      Guu, Kelvin",
    booktitle = "Proceedings of the 2021 Conference on Empirical Methods in Natural Language Processing",
    month = nov,
    year = "2021",
    address = "Online and Punta Cana, Dominican Republic",
    publisher = "Association for Computational Linguistics",
    doi = "10.18653/v1/2021.emnlp-main.607",
    pages = "7683--7698",
}

@inproceedings{wang-etal-2022-training,
    title = "Training Data is More Valuable than You Think: A Simple and Effective Method by Retrieving from Training Data",
    author = "Wang, Shuohang  and
      Xu, Yichong  and
      Fang, Yuwei  and
      Liu, Yang  and
      Sun, Siqi  and
      Xu, Ruochen  and
      Zhu, Chenguang  and
      Zeng, Michael",
    booktitle = "Proceedings of the 60th Annual Meeting of the Association for Computational Linguistics (Volume 1: Long Papers)",
    month = may,
    year = "2022",
    address = "Dublin, Ireland",
    publisher = "Association for Computational Linguistics",
    doi = "10.18653/v1/2022.acl-long.226",
    pages = "3170--3179",
}

@INPROCEEDINGS{nashid2023retrieval,
  author={Nashid, Noor and Sintaha, Mifta and Mesbah, Ali},
  booktitle={2023 IEEE/ACM 45th International Conference on Software Engineering (ICSE)}, 
  title={Retrieval-Based Prompt Selection for Code-Related Few-Shot Learning}, 
  year={2023},
  volume={},
  number={},
  pages={2450-2462},
  doi={10.1109/ICSE48619.2023.00205}
}

@inproceedings{agrawal-etal-2023-context,
    title = "In-context Examples Selection for Machine Translation",
    author = "Agrawal, Sweta  and
      Zhou, Chunting  and
      Lewis, Mike  and
      Zettlemoyer, Luke  and
      Ghazvininejad, Marjan",
    booktitle = "Findings of the Association for Computational Linguistics: ACL 2023",
    month = jul,
    year = "2023",
    address = "Toronto, Canada",
    publisher = "Association for Computational Linguistics",
    doi = "10.18653/v1/2023.findings-acl.564",
    pages = "8857--8873",
}

@inproceedings{gao-etal-2021-making,
    title = "Making Pre-trained Language Models Better Few-shot Learners",
    author = "Gao, Tianyu  and
      Fisch, Adam  and
      Chen, Danqi",
    booktitle = "Proceedings of the 59th Annual Meeting of the Association for Computational Linguistics and the 11th International Joint Conference on Natural Language Processing (Volume 1: Long Papers)",
    month = aug,
    year = "2021",
    address = "Online",
    publisher = "Association for Computational Linguistics",
    doi = "10.18653/v1/2021.acl-long.295",
    pages = "3816--3830",
}

@article{koksal2022meal,
  title={MEAL: Stable and Active Learning for Few-Shot Prompting},
  author={K{\"o}ksal, Abdullatif and Schick, Timo and Sch{\"u}tze, Hinrich},
  journal={arXiv preprint arXiv:2211.08358},
  year={2022}
}

@inproceedings{schroder-etal-2022-revisiting,
    title = "Revisiting Uncertainty-based Query Strategies for Active Learning with Transformers",
    author = {Schr{\"o}der, Christopher  and
      Niekler, Andreas  and
      Potthast, Martin},
    booktitle = "Findings of the Association for Computational Linguistics: ACL 2022",
    month = may,
    year = "2022",
    address = "Dublin, Ireland",
    publisher = "Association for Computational Linguistics",
    doi = "10.18653/v1/2022.findings-acl.172",
    pages = "2194--2203",
}

@inproceedings{shin-etal-2021-constrained,
    title = "Constrained Language Models Yield Few-Shot Semantic Parsers",
    author = "Shin, Richard  and
      Lin, Christopher  and
      Thomson, Sam  and
      Chen, Charles  and
      Roy, Subhro  and
      Platanios, Emmanouil Antonios  and
      Pauls, Adam  and
      Klein, Dan  and
      Eisner, Jason  and
      Van Durme, Benjamin",
    booktitle = "Proceedings of the 2021 Conference on Empirical Methods in Natural Language Processing",
    month = nov,
    year = "2021",
    address = "Online and Punta Cana, Dominican Republic",
    publisher = "Association for Computational Linguistics",
    doi = "10.18653/v1/2021.emnlp-main.608",
    pages = "7699--7715",
}

@inproceedings{maronikolakis-etal-2023-improving,
    title = "Improving Few-Shot Learning with Multilingual Transfer and {M}onte {C}arlo Training Set Selection",
    author = {Maronikolakis, Antonis  and
      O{'}Grady, Paul  and
      Sch{\"u}tze, Hinrich  and
      Lyra, Matti},
    booktitle = "Proceedings of the 2023 CLASP Conference on Learning with Small Data (LSD)",
    month = sep,
    year = "2023",
    address = "Gothenburg, Sweden",
    publisher = "Association for Computational Linguistics",
    pages = "1--10",
}

@article{nguyen2023context,
  title={In-context Example Selection with Influences},
  author={Nguyen, Tai and Wong, Eric},
  journal={arXiv preprint arXiv:2302.11042},
  year={2023}
}

@article{agarwal2021sensitivity,
  title={On sensitivity of meta-learning to support data},
  author={Agarwal, Mayank and Yurochkin, Mikhail and Sun, Yuekai},
  journal={Advances in Neural Information Processing Systems},
  volume={34},
  pages={20447--20460},
  year={2021}
}

@inproceedings{su2022selective,
  title={Selective Annotation Makes Language Models Better Few-Shot Learners},
  author={Su, Hongjin and Kasai, Jungo and Wu, Chen Henry and Shi, Weijia and Wang, Tianlu and Xin, Jiayi and Zhang, Rui and Ostendorf, Mari and Zettlemoyer, Luke and Smith, Noah A and others},
  booktitle={The Eleventh International Conference on Learning Representations},
  year={2022}
}

@inproceedings{levy-etal-2023-diverse,
    title = "Diverse Demonstrations Improve In-context Compositional Generalization",
    author = "Levy, Itay  and
      Bogin, Ben  and
      Berant, Jonathan",
    booktitle = "Proceedings of the 61st Annual Meeting of the Association for Computational Linguistics (Volume 1: Long Papers)",
    month = jul,
    year = "2023",
    address = "Toronto, Canada",
    publisher = "ACL",
    doi = "10.18653/v1/2023.acl-long.78",
    pages = "1401--1422",
}

@inproceedings{ye-etal-2023-complementary,
    title = "Complementary Explanations for Effective In-Context Learning",
    author = "Ye, Xi  and
      Iyer, Srinivasan  and
      Celikyilmaz, Asli  and
      Stoyanov, Veselin  and
      Durrett, Greg  and
      Pasunuru, Ramakanth",
    booktitle = "Findings of the Association for Computational Linguistics: ACL 2023",
    month = jul,
    year = "2023",
    address = "Toronto, Canada",
    publisher = "Association for Computational Linguistics",
    doi = "10.18653/v1/2023.findings-acl.273",
    pages = "4469--4484",
}

@inproceedings{rubin-etal-2022-learning,
    title = "Learning To Retrieve Prompts for In-Context Learning",
    author = "Rubin, Ohad  and
      Herzig, Jonathan  and
      Berant, Jonathan",
    booktitle = "Proceedings of the 2022 Conference of the North American Chapter of the Association for Computational Linguistics: Human Language Technologies",
    month = jul,
    year = "2022",
    address = "Seattle, United States",
    publisher = "ACL",
    doi = "10.18653/v1/2022.naacl-main.191",
    pages = "2655--2671",
}

@misc{wang2023learning,
      title={Learning to Sample Tasks for Meta Learning}, 
      author={Jingyao Wang and Zeen Song and Xingzhe Su and Lingyu Si and Hongwei Dong and Wenwen Qiang and Changwen Zheng},
      year={2023},
      eprint={2307.08924},
      archivePrefix={arXiv},
      primaryClass={cs.LG}
}

@article{aimen2023leveraging,
  title={Leveraging Task Variability in Meta-learning},
  author={Aimen, Aroof and Ladrecha, Bharat and Sidheekh, Sahil and Krishnan, Narayanan C},
  journal={SN Computer Science},
  volume={4},
  number={5},
  pages={539},
  year={2023},
  publisher={Springer}
}

@inproceedings{jundi-lapesa-2022-translate,
    title = "How to Translate Your Samples and Choose Your Shots? Analyzing Translate-train {\&} Few-shot Cross-lingual Transfer",
    author = "Jundi, Iman  and
      Lapesa, Gabriella",
    booktitle = "Findings of the Association for Computational Linguistics: NAACL 2022",
    month = jul,
    year = "2022",
    address = "Seattle, United States",
    publisher = "Association for Computational Linguistics",
    doi = "10.18653/v1/2022.findings-naacl.11",
    pages = "129--150",
}

@article{luo2023dr,
  title={Dr. ICL: Demonstration-Retrieved In-context Learning},
  author={Luo, Man and Xu, Xin and Dai, Zhuyun and Pasupat, Panupong and Kazemi, Mehran and Baral, Chitta and Imbrasaite, Vaiva and Zhao, Vincent Y},
  journal={arXiv preprint arXiv:2305.14128},
  year={2023}
}

@article{scarlatos2023reticl,
  title={RetICL: Sequential Retrieval of In-Context Examples with Reinforcement Learning},
  author={Scarlatos, Alexander and Lan, Andrew},
  journal={arXiv preprint arXiv:2305.14502},
  year={2023}
}

@article{song2023survey,
author = {Song, Yisheng and Wang, Ting and Cai, Puyu and Mondal, Subrota K. and Sahoo, Jyoti Prakash},
title = {A Comprehensive Survey of Few-Shot Learning: Evolution, Applications, Challenges, and Opportunities},
year = {2023},
issue_date = {December 2023},
publisher = {Association for Computing Machinery},
address = {New York, NY, USA},
volume = {55},
number = {13s},
issn = {0360-0300},
doi = {10.1145/3582688},
journal = {ACM Comput. Surv.},
month = {jul},
articleno = {271},
numpages = {40}
}

@article{hospedales2021meta,
  title={Meta-learning in neural networks: A survey},
  author={Hospedales, Timothy and Antoniou, Antreas and Micaelli, Paul and Storkey, Amos},
  journal={IEEE transactions on pattern analysis and machine intelligence},
  volume={44},
  number={9},
  pages={5149--5169},
  year={2021},
  publisher={IEEE}
}

@article{vanschoren2018meta,
  title={Meta-learning: A survey},
  author={Vanschoren, Joaquin},
  journal={arXiv preprint arXiv:1810.03548},
  year={2018}
}

@article{dong2022survey,
  title={A survey for in-context learning},
  author={Dong, Qingxiu and Li, Lei and Dai, Damai and Zheng, Ce and Wu, Zhiyong and Chang, Baobao and Sun, Xu and Xu, Jingjing and Sui, Zhifang},
  journal={arXiv preprint arXiv:2301.00234},
  year={2022}
}

@article{liu2023pre,
author = {Liu, Pengfei and Yuan, Weizhe and Fu, Jinlan and Jiang, Zhengbao and Hayashi, Hiroaki and Neubig, Graham},
title = {Pre-Train, Prompt, and Predict: A Systematic Survey of Prompting Methods in Natural Language Processing},
year = {2023},
issue_date = {September 2023},
publisher = {Association for Computing Machinery},
address = {New York, NY, USA},
volume = {55},
number = {9},
issn = {0360-0300},
doi = {10.1145/3560815},
journal = {ACM Comput. Surv.},
month = {jan},
articleno = {195},
numpages = {35}
}

@article{tibshirani1996regression,
  title={Regression shrinkage and selection via the lasso},
  author={Tibshirani, Robert},
  journal={Journal of the Royal Statistical Society Series B: Statistical Methodology},
  volume={58},
  number={1},
  pages={267--288},
  year={1996},
  publisher={Oxford University Press}
}

@inproceedings{he2016deep,
  title={Deep residual learning for image recognition},
  author={He, Kaiming and Zhang, Xiangyu and Ren, Shaoqing and Sun, Jian},
  booktitle={Proceedings of the IEEE conference on computer vision and pattern recognition},
  pages={770--778},
  year={2016}
}

@inproceedings{deng2009imagenet,
  title={Imagenet: A large-scale hierarchical image database},
  author={Deng, Jia and Dong, Wei and Socher, Richard and Li, Li-Jia and Li, Kai and Fei-Fei, Li},
  booktitle={2009 IEEE conference on computer vision and pattern recognition},
  pages={248--255},
  year={2009},
  organization={Ieee}
}

@inproceedings{devlin-etal-2019-bert,
    title = "{BERT}: Pre-training of Deep Bidirectional Transformers for Language Understanding",
    author = "Devlin, Jacob  and
      Chang, Ming-Wei  and
      Lee, Kenton  and
      Toutanova, Kristina",
    booktitle = "Proceedings of the 2019 Conference of the North {A}merican Chapter of the ACL: Human Language Technologies",
    month = jun,
    year = "2019",
    address = "Minneapolis, Minnesota",
    publisher = "ACL",
    doi = "10.18653/v1/N19-1423",
    pages = "4171--4186",
}

@article{zhou2024lima,
  title={Lima: Less is more for alignment},
  author={Zhou, Chunting and Liu, Pengfei and Xu, Puxin and Iyer, Srinivasan and Sun, Jiao and Mao, Yuning and Ma, Xuezhe and Efrat, Avia and Yu, Ping and Yu, Lili and others},
  journal={Advances in Neural Information Processing Systems},
  volume={36},
  year={2024}
}

@article{grattafiori2024llama,
  title={The llama 3 herd of models},
  author={Grattafiori, Aaron and Dubey, Abhimanyu and Jauhri, Abhinav and Pandey, Abhinav and Kadian, Abhishek and Al-Dahle, Ahmad and Letman, Aiesha and Mathur, Akhil and Schelten, Alan and Vaughan, Alex and others},
  journal={arXiv preprint arXiv:2407.21783},
  year={2024}
}

@article{gemma_2025,
    title={Gemma 3},
    url={https://goo.gle/Gemma3Report},
    publisher={Kaggle},
    author={Gemma Team},
    year={2025}
}

@misc{qwen2.5,
    title = {Qwen2.5: A Party of Foundation Models},
    url = {https://qwenlm.github.io/blog/qwen2.5/},
    author = {Qwen Team},
    month = {September},
    year = {2024}
}

@inproceedings{diao-etal-2024-active,
    title = "Active Prompting with Chain-of-Thought for Large Language Models",
    author = "Diao, Shizhe  and
      Wang, Pengcheng  and
      Lin, Yong  and
      Pan, Rui  and
      Liu, Xiang  and
      Zhang, Tong",
    editor = "Ku, Lun-Wei  and
      Martins, Andre  and
      Srikumar, Vivek",
    booktitle = "Proceedings of the 62nd Annual Meeting of the Association for Computational Linguistics (Volume 1: Long Papers)",
    month = aug,
    year = "2024",
    address = "Bangkok, Thailand",
    publisher = "Association for Computational Linguistics",
    url = "https://aclanthology.org/2024.acl-long.73/",
    doi = "10.18653/v1/2024.acl-long.73",
    pages = "1330--1350"
}

@inproceedings{purohit-etal-2024-explora,
    title = "{EXPLORA}: Efficient Exemplar Subset Selection for Complex Reasoning",
    author = "Purohit, Kiran  and
      V, Venktesh  and
      Devalla, Raghuram  and
      Yerragorla, Krishna Mohan  and
      Bhattacharya, Sourangshu  and
      Anand, Avishek",
    editor = "Al-Onaizan, Yaser  and
      Bansal, Mohit  and
      Chen, Yun-Nung",
    booktitle = "Proceedings of the 2024 Conference on Empirical Methods in Natural Language Processing",
    month = nov,
    year = "2024",
    address = "Miami, Florida, USA",
    publisher = "Association for Computational Linguistics",
    url = "https://aclanthology.org/2024.emnlp-main.307/",
    doi = "10.18653/v1/2024.emnlp-main.307",
    pages = "5367--5388"
}

@inproceedings{purohit2025sample,
title={Sample Efficient Demonstration Selection for In-Context Learning},
author={Kiran Purohit and Venktesh V and Sourangshu Bhattacharya and Avishek Anand},
booktitle={Forty-second International Conference on Machine Learning},
year={2025},
url={https://openreview.net/forum?id=cuqvlLBQK6}
}

@article{
albalak2024a,
title={A Survey on Data Selection for Language Models},
author={Alon Albalak and Yanai Elazar and Sang Michael Xie and Shayne Longpre and Nathan Lambert and Xinyi Wang and Niklas Muennighoff and Bairu Hou and Liangming Pan and Haewon Jeong and Colin Raffel and Shiyu Chang and Tatsunori Hashimoto and William Yang Wang},
journal={Transactions on Machine Learning Research},
issn={2835-8856},
year={2024},
url={https://openreview.net/forum?id=XfHWcNTSHp},
note={Survey Certification, Featured Certification}
}

@inproceedings{cegin-etal-2025-use,
    title = "Use Random Selection for Now: Investigation of Few-Shot Selection Strategies in {LLM}-based Text Augmentation",
    author = "Cegin, Jan  and
      Pecher, Branislav  and
      Simko, Jakub  and
      Srba, Ivan  and
      Bielikova, Maria  and
      Brusilovsky, Peter",
    editor = "Christodoulopoulos, Christos  and
      Chakraborty, Tanmoy  and
      Rose, Carolyn  and
      Peng, Violet",
    booktitle = "Findings of the Association for Computational Linguistics: EMNLP 2025",
    month = nov,
    year = "2025",
    address = "Suzhou, China",
    publisher = "Association for Computational Linguistics",
    url = "https://aclanthology.org/2025.findings-emnlp.296/",
    doi = "10.18653/v1/2025.findings-emnlp.296",
    pages = "5533--5550",
    ISBN = "979-8-89176-335-7"
}

\appendix

\section{ACSESS Method Supplementary Details}
\label{app:acsess-details}

In this appendix, we provide and discuss supplementary details of our proposed strategy. This includes detailed steps of the forward selection in Algorithm~\ref{alg:forward}, backward selection in Algorithm~\ref{alg:backward} and datamodels selection in Algorithm~\ref{alg:datamodels}. In addition, we provide low-level details how the method works, suggestions for using the method in order to consider its generalisability and reducing its model/dataset dependence, the computational costs of the method (in comparison with other baselines) and how we reduce them to provide compute optimised version of our method, and, finally, suggestions for selecting the best configuration of the ACSESS method based on the compute-performance trade-off of these configurations. Some parts of this section are based on the observations from our experiments.

\begin{algorithm}[!tbh]
\caption{Forward Selection} \label{alg:forward}
\begin{algorithmic}[1]
\small
\Require $N$ - number of samples to choose
\Require $\mathbf{S}$ - set of sample selection strategies
\State $\mathbf{S_{F}} \gets \{\}$ 
\State $best\_perf$ = -1
\While{Performance is increasing \textbf{or} $\mathbf{S_F}$ is empty}
    \ForAll{$s$ in $\mathbf{S}$}
        \State Create $\mathbf{S_{temp}}$ by combining $\mathbf{S_{F}}$ and $s$
        \State Set $w_s=1/|\mathbf{S_{temp}}|$
        \State Calculate for each sample $score_{temp}(x) = \sum_{s \in S_{temp}} w_s * objective_s(x)$
        \State Calculate $perf_i$ by evaluating the few-shot model on $N$ samples selected using $score_{temp}$
        \If{$perf_{i} > best\_perf$}
            \State $best\_perf = perf_{i}$
            \State $strategy$ = $s$
        \EndIf
    \EndFor
    \If{Any $strategy$ improves performance}
        \State Add found strategy to $\mathbf{S_{F}}$
        \State Remove found strategy from $\mathbf{S}$
    \EndIf
\EndWhile
\State \Return $\mathbf{S_{F}}$
\end{algorithmic}
\end{algorithm}

\begin{algorithm}[!tbh]
\caption{Backward Selection} \label{alg:backward}
\begin{algorithmic}[1]
\small
\Require $N$ - number of samples to choose
\Require $\mathbf{S}$ - set of sample selection strategies

\State $\mathbf{S_{B}}$ = $\mathbf{S}$
\State Set $w=1/|\mathbf{S_{B}}|$
\State Calculate for each sample $score_{temp}(x) = \sum_{i}^{|\mathbf{S_{B}}|} w_s * objective_s(x)$
\State Calculate $best\_perf$ by evaluating the few-shot model on $N$ samples selected using $score_{temp}$
\While{Performance is increasing}
    \ForAll{$s_{i}$ in $\mathbf{S}$}
        \State Create $\mathbf{S_{temp}}$ by removing $s_{i}$ from $\mathbf{S_{B}}$
        \State Set $w_s=1/|\mathbf{S_{B}}|$
        \State Calculate for each sample $score_{temp}(x) = \sum_{s \in S_{B}} w_s * objective_s(X)$
        \State Calculate $perf_i$ by evaluating the few-shot model on $N$ samples selected using $score_{temp}$
        \If{$perf_{i} > best\_perf$}
            \State $best\_perf = perf_{i}$
            \State $strategy$ = $s$
        \EndIf
    \EndFor
    \If{Any $strategy$ improves performance}
        \State Remove found strategy from $\mathbf{S_{B}}$
    \EndIf
\EndWhile
\State \Return $\mathbf{S_{B}}$
\end{algorithmic}
\end{algorithm}

\begin{algorithm}[tbh]
\caption{Datamodels Selection} \label{alg:datamodels}
\begin{algorithmic}[1]
\small
\Require $N$ - number of samples to choose
\Require $\mathbf{S}$ - set of sample selection strategies
\Require $perf$ - performance of the few-shot model on the baseline setting

\State $\mathbf{S_{D}} \gets \{\}$
\State Determine number of random combinations to create $M$
\State Create a set $\mathbf{C}$ containing $M$ unique, random combinations of strategies from $\mathbf{S}$, ensuring each strategy is contained at least 5 times
\ForAll{$C_{j}$ in $\mathbf{C}$}
    \State Set $w_s=1/|\mathbf{C_{j}}|$
    \State Calculate for each sample $score_{j}(x) = \sum_{s \in C_j} w_s * objective_s(x)$
    \State Calculate $perf_j$ by evaluating the few-shot model on $N$ samples selected using $scores_{j}$
    \State Calculate $diff_{j}$ = $perf_{j}$ - $perf$
    \State Add $diff_{j}$ to the list of differences to baseline $Diff$
    \State Create a new multi-label presence vector using strategies from $C_{j}$ and add it to the list $Strategies$
\EndFor
\State Train a $LASSO$ linear regression using $Strategies$ as features and $Diff$ as targets
\State Obtain vector of weights $\vec{W}$, each corresponding to one strategy in the multi-label presence vector
\State Select strategies with positive weights in $\vec{W}$ and add them to $\mathbf{S_{D}}$
\State Set $W_D$ using weights of the selected strategies from $\mathbf{S_{D}}$
\State \Return $\mathbf{S_{D}}$, ${W}_{D}$
\end{algorithmic}
\end{algorithm}

\paragraph{Considerations for the computation costs of the ACSESS method} Although the proposed ACSESS method may at first seem to require significantly more computation resources than the single-property selection strategies or the in-context learning specific baselines, there are easy to apply considerations and modifications that increase the compute-efficiency and reduce the costs. We define this analysis into two separate parts as they introduce different computation costs: 1) using the method in practice after the optimal set of strategies is already identified; and 2) identifying the optimal set of strategies to combine. As all of the commonly used single-property strategies perform the selection by training the model for a few epochs (or just running an inference in order to obtain embeddings in case of Similarity and Diversity) and then using the training dynamics, these strategies can be run at the same time, reusing the same trained model. As such, when using the proposed ACSESS method in practice, we need to train the model only a single time and can use it across all the combined strategies. Therefore the cost in this case is equivalent to the single-property selection strategies (and lower than the in-context learning specific baselines). Similar optimisation can also be used when identifying the optimal set of strategies to combine, where the training of the model is the main computation cost. Even though the strategies are combined with different ones, they are designed to assign a specific score to each sample, which can then be reused in the combination. In other words, the strategies are needed to be run only once to obtain the scores which are then used as strong approximation for the objectives. The only cost that remains is the cost of evaluating the set of selected strategies by each of the combination. In case of the gradient-based few-shot learning strategies, this cost is often negligible, as we are using only a small number of samples. However, in the case of in-context learning, the cost scales with the size of the model used and the number of combinations that are explored, as the inference itself is quite expensive. The cost of this evaluation step is also dependent on the number of samples and the model used (i.e., training on 25 samples vs. full dataset; inference on smaller vs. larger models). Due to this, the computation cost for the ACSESS strategy is higher (by the number of required inferences and the cost of single inference) over the single-property selection strategy. At the same time, the cost is lower or on par with in-context learning specific baselines, which use multiple inferences or trainings of the model while selecting the samples (even though such comparison is hard to make and is based only on our empirical observations during the experiments). Finally, the computation cost can be further reduced by foregoing the model and dataset specific tuning (e.g., using the ACSESS method on multiple datasets at the same time), as described in the following section. The specific cost comparison of the ACSESS with the baselines is presented in Table~\ref{tab:gpu-costs}. The costs are calculated as the average GPU wall-clock time for each strategy averaged across all datasets.

\begin{table}[tbh]
\centering
\small
\caption{Comparison of costs, calculated as the average GPU wall-clock time across all datasets, for the individual sample selection strategies. The ACSESS strategy performs on par with in-context learning specific strategies, while showing higher cost than single property selection strategies.}
\label{tab:gpu-costs}
\begin{tabular}{lc}
\textsc{Strategy}             & \textsc{GPU time}  \\ \toprule
Similarity/Diversity & 0.56      \\
Active Learning      & 1.02-1.11 \\
Core-Set             & 0.97-1.24 \\ \midrule
LENS                 & 6.19      \\
Active Prompt        & 5.37      \\
EXPLORA              & 4.98      \\
CASE                 & 2.04      \\ \midrule
ACSESS\textsubscript{Weighted}               & 2.05      \\
ACSESS\textsubscript{Uniform}     & 1.48      \\

\bottomrule
\end{tabular}

\end{table}

\paragraph{Considerations for using the ACSESS method and reducing its model/dataset dependence} Similarly to hyperparameter optimisation, we suggest to identify and combine the different strategies using the validation set to prevent any data leakage (i.e., evaluating the performance of different strategies using the validation set). At its core, the method is designed to be run separately for each model and dataset. Although such setting leads to the most benefit and highest increase in performance, this makes it significantly model/dataset dependent and requires significantly more computation resources. In order to reduce the model/dataset dependence, the different steps of the method be run across multiple models or datasets. This may be best illustrated on the identification of strategies and further weighting using the Datamodels method. Instead of having a separate set of combinations of strategies for each dataset a model, we can concatenate them together to create a single dataset. After training the LASSO method on this concatenated dataset, we obtain the relevant strategies and their weights for all the dataset and models. Similar modification can be applied to the forward and backward selection as well, where we evaluate the added/removed strategy on all the models/datasets, aggregate the score and select the strategies on this aggregated score. However, as we observed a strong dependence on the few-shot approach and the data modality, we suggest to run the ACSESS method separately for each few-shot learning approach and modality. In our experiments we also follow this suggestion and aggregate only over the datasets, as we have noticed only negligible changes when running the ACSESS method separately for each dataset (i.e., in some cases, the set of identified strategies we report in the results included 1 more strategy that performed really well on that specific dataset). Running the ACSESS strategy on even higher aggregation (over all the approaches, models and modalities) is not recommended as our preliminary experiments indicate that it would not lead to significant performance benefit and the method could fail to identify any relevant selection strategy in some cases. Finally, foregoing the specific model/dataset weighted combination and instead using the uniform combination (or the weights provided by the Datamodels approach across multiple datasets) can increase the transferability and generalisability of the ACSESS method further.

\paragraph{Choosing the best configuration for the ACSESS method} Based on the results of our experiments, to get the best performance benefit of the ACSESS method, running the full strategy identification, either separately for each model and dataset or aggregated across datasets, and using the weighted combination of these strategies should be used. This configuration provides the best results across all datasets, approaches and models, with the exception of Few-Shot Fine-Tuning. However, this may require extensive computation resources. If such resources are not available, or if we want to quickly determine the potential benefit of the sample selection, using a uniform combination of the \textit{Cartography} and \textit{Margin} (and in some cases \textit{Forgetting}) selection strategies provides a good approximation. This configuration does not incur any additional computation costs, while providing consistent improvement on all the datasets and models, which is only slightly lower than when using the model/dataset specific configuration. Although, running the full method should always provide the most benefit, this configuration (uniform combination of the 3 best performing strategies identified in our experiments) is a good alternative that is transferable across all settings.

\section{Sample Selection Strategies Details}
\label{app:sss-details}

\begin{table*}[!tbh]
\centering
\small
\caption{Common single-property selection strategies used in the ACSESS method, with the description of their objective, and categorised based on the sample property they consider and the group of selection strategies it belongs to.}
\label{tab:investigated-strategies}

\begin{tabularx}{\textwidth}{@{}m{0.02cm}p{0.125\textwidth}p{0.15\textwidth}Xp{0.215\textwidth}@{}}

\toprule
& \textbf{Group} & \textbf{Strategy} & \textbf{Description} & \textbf{Reference} \\ \midrule


\multirow[c]{14}{*}{\rotatebox{90}{\begin{footnotesize}\textbf{Informativeness}\end{footnotesize}}} 
& Similarity & Similarity & Most similar samples. & --- \\
& Similarity & Diversity & Most diverse/dissimilar samples. & --- \\
& Active Learn. & Entropy& Highest entropy over class probabilities. & \citep{park2022active}  \\
& Active Learn. & Margin & Lowest difference in class probability between the 2 most probable classes. & \citep{park2022active}  \\
& Active Learn. & Least Confidence & Class with lowest assigned probability. & \citep{park2022active}  \\
& Active Learn. & Loss & Highest loss. & \citep{park2022active}  \\
& Core-Set & Contrastive Active Learning (CAL) & Predictive likelihood that diverges most from the neighbourhood as determined by KL divergence. & \citep{margatina-etal-2021-active} \\
& Core-Set & DeepFool & Smallest amount of perturbation to change class. & \citep{ducoffe2018adversarial} \\
& Core-Set & GraNd & Highest contribution to decline of training loss. & \citep{paul2021grand} \\
& Core-Set & Graph-Cut & Submodularity function that measures diversity and informativeness. & \citep{iyer2013submodular, iyer2021submodular} \\ \midrule

\multirow[c]{7}{*}{\rotatebox{90}{\begin{footnotesize}\textbf{Representativeness}\end{footnotesize}}} 
& Core-Set & Herding & Minimising distance of subset centre to full dataset. & \citep{welling2009herding, chen2010herding} \\
& Core-Set & KCenter & Minimising distance of every sample in subset to full dataset. & \citep{sener2018active, agarwal2020contextual} \\
& Core-Set & CRAIG & Gradients representative of full dataset. & \citep{mirzasoleiman2020coresets} \\
& Core-Set & Glister & Bi-level optimisation maximising log-likelihood. & \citep{killamsetty2021glister} \vspace{0.2em}\\ \midrule

\multirow[c]{4}{*}{\rotatebox{90}{\begin{footnotesize}\textbf{Learnability}\end{footnotesize}}} 
& Core-Set & Forgetting & How often the samples are forgotten. & \citep{toneva2018empirical} \\
& Cartography & Cartography & How easy it is to learn the samples. Due to no obvious consensus, all of easy, ambiguous, hard and combination of easy and ambiguous samples are considered. & \citep{swayamdipta-etal-2020-dataset, zhang-plank-2021-cartography-active} \\
\bottomrule
\end{tabularx}
\end{table*}

In this appendix, we provide further details about the different single-property sample selection strategies that we evaluate in this paper, as well as for the baselines. In addition, we specify the hyperparameters of these methods and how we use them for selecting the set of samples. The high-level overview of the single-property strategies is included in Table~\ref{tab:investigated-strategies}.

The idea behind the \textbf{uncertainty-based} and \textbf{active learning} selection strategies is that the most informative samples with higher impact are the ones that the model is least certain about~\citep{coleman2020Selection, park2022active}. These strategies select samples iteratively over multiple steps. In our experiments, we select 1 sample for every class in individual step. In addition, the first sample for every class is chosen randomly, as such setting was observed to lead to better active learning results~\citep{park2022active}. When using these methods for combination, the samples that are selected after running all the required steps (e.g., 4 steps when choosing 5-shots) use the score assigned to them at their respective step when they were chosen, while the remaining samples are assigned the average score over all the selection steps. The \textbf{Entropy} strategy measures the entropy for each of the samples over the predicted class probabilities (which is used as the sample score) and selects the sample with highest entropy. The \textbf{Margin (Breaking Ties)} strategy assigns to each sample the difference between the probability of the two most probable classes. The samples with lowest such difference are selected as those are the ones the model is most uncertain about (and which can be considered on the decision boundary of the model). The \textbf{Least Confidence} strategy simply selects the samples where the most likely class has the lowest probability assigned to it. Finally, the \textbf{Loss} strategy selects samples with the highest loss. 

The idea behind the core-set selection strategies is to select subset of samples that are representative of the whole dataset (i.e., the subset has similar properties to the full set). The included strategies select samples based on different criteria, such as error/loss, gradient or submodularity. These strategies select the full set of samples at the same time according to the assigned scores. As such, when using these strategies for combination, the full sample scores are used from this single iteration. 

The \textbf{Decision Boundary} based methods select samples the are close to the decision boundary, as these represent the samples that are hard to separate and should provide the most information to the models. The \textbf{Contrastive Active Learning (CAL)}~\citep{margatina-etal-2021-active} selects the samples whose predictive likelihood diverges the most from their neighbourhood as determined by the KL divergence. The \textbf{DeepFool} strategy~\citep{ducoffe2018adversarial} approximates the samples on boundary by perturbing the samples until the predicted class changes and selects the samples that require the smallest perturbation. 

The idea behind \textbf{Error or Loss} based methods is that the samples that contribute more loss or error to the training of model have higher importance and should contribute more to the training when selected. The \textbf{GraNd} strategy~\citep{paul2021grand} calculates the average contribution of each sample to the decline of the training loss at early stage of training across several independent training runs (calculating gradient norm expectation of the sample). The \textbf{Forgetting} strategy~\citep{toneva2018empirical} simply calculates how often the specific sample is incorrectly classified after being classified correctly in the previous epoch. The number of such forgetting events is used as the sample score and samples that are forgotten the most often (or least often) are selected. In our evaluation of single-property strategies, we consider only the setting of most often forgotten samples (as it has shown the best performance in the preliminary experiments). However, when using the strategy for combination, we also consider the setting of selecting the least forgotten ones. 

The \textbf{Submodularity} based methods use submodular functions~\citep{iyer2013submodular, iyer2021submodular}, which naturally measure the informativeness and diversity of a subset of samples. We evaluate the selection based on \textbf{Graph-Cut}, as the method performed the best from all the different submodular functions in the preliminary experiments. 

The \textbf{Geometry} based methods assume that samples that are close in the feature space have similar properties. As such, these methods try to remove samples that are close to each other, as they provide redundant information. Removing enough of such samples then leads to a small subset of sample that cover the properties of the whole dataset. The \textbf{Herding} strategy~\citep{welling2009herding, chen2010herding} iteratively and greedily adds one sample to the subset which minimises the distance between the centre of the subset and the centre of the whole dataset. On the other hand, the \textbf{KCenter} strategy~\citep{sener2018active, agarwal2020contextual} selects sample that minimises the largest distance between each sample already in the subset and its closest sample not yet included in subset. 

The \textbf{Gradient Matching} based methods try to find a subset of samples whose gradient can imitate the whole dataset. The \textbf{CRAIG} strategy~\citep{mirzasoleiman2020coresets} converts the problem into maximisation of a monotone submodular function which is then solved greedily. 

Finally, the \textbf{Bilevel Optimisation} based methods convert the problem into a bilevel optimisation, with different objectives in the outer and inner level. The \textbf{Glister} strategy~\citep{killamsetty2021glister} uses a validation set on the outer optimisation, where the subset selection is used as objective, and log-likelihood in the inner optimisation, which handles model parameter optimisation.

In addition, we also evaluate and use sample selection based on the \textbf{Cartography}~\citep{swayamdipta-etal-2020-dataset, zhang-plank-2021-cartography-active}, which measures how easy or hard it is to learn different samples. This ease of learning is determined by observing the training dynamics of all the samples across few epochs, looking at the average confidence/probability of the correct class and the variance of this confidence. The samples with high confidence and low variance are considered to be the \textit{easy} to learn samples. At the same time the samples with small confidence and small to medium variance are considered the \textit{hard} to learn ones. The remaining samples are considered to be \textit{ambiguous} (medium confidence or samples with high variance). We explore four different settings in our experiments and choose the samples accordingly: 1) \textbf{Easy} samples, where we sort the samples based on confidence and choose the top-K samples with highest confidence for each class; 2) \textbf{Hard} samples, where we sort the samples based on confidence and choose the bottom-K samples for each class (i.e., the lowest confidence samples); 3) \textbf{Ambiguous}, where we first calculate average confidence and standard deviation of all the samples, select the samples whose confidence is close around the average confidence (in the interval defined by 1 standard deviation around the average confidence), and then randomly sample from them; and 4) \textbf{Easy + Ambiguous}, where we choose half of the samples from the \textit{easy} set and the other half from \textit{ambiguous} set.

All the approaches mentioned so far perform the selection using a model that is trained for some epochs. The number of epochs the training is done for significantly affects the outcome of different strategies. For this reason, we search for an optimal number of epochs by evaluating the selection strategies and the samples they select on a single run. The optimal number of epochs determined by this search is 10\% of the full epochs used for the few-shot fine-tuning models. This number represents a good trade-off between the computational cost of the methods and the quality of samples. Running the training for full number of epochs may result in better selection of samples, but we consider such setting to not be representative. In addition, we use the same number of epochs for each strategy to keep the comparison as representative as possible. In addition, we also search for an optimal set of hyperparameters for the selection strategies (such as the learning rate, the optimiser used or batch size). The optimal set of hyperparameters that are used throughout the experiments is the use of the Adam optimiser with Cosine Annealing learning rate scheduler, with learning rate of 0.01 for image data and 0.0001 for text data and a batch size of 64.

To keep the results as comparable as possible, each selection strategy uses the same base model as the one used in the main experiments, i.e., ResNet-18~\citep{he2016deep} pretrained on Imagenet dataset~\citep{deng2009imagenet} for image data and pretrained BERT-base~\citep{devlin-etal-2019-bert} for text data in case of gradient few-shot learning, and the same large language models (Mistral~\citep{jiang2023mistral} and Zephyr~\citep{tunstall2023zephyr}), but with only their last classification layer being learned, for in-context learning. In case the selection strategy requires sample features, we use the output of last fully-connected layer (the input of classification layer) as the feature vector similar to~\citet{guo2022deepcore}.

We also evaluate the sample selection based on \textbf{Similarity} and \textbf{Diversity} as it represents popular approach for in-context learning. To select the samples, we use the cosine similarity between the feature representation (obtained from the penultimate layer) of the samples. For in-context learning, we instead the most similar (for Similarity) or the most dissimilar (for Diversity) samples for each test sample as is common for this approach~\citep{sun2023pushing}. To select the samples for gradient few-shot learning, we first randomly select one sample for each class. Afterwards, we expand the set of selected samples by iteratively adding the most similar (for Similarity) or most dissimilar (for Diversity) sample to the average feature representation of the set of selected samples for each class (i.e., if we already have 5 selected samples in the specific class, we calculate the similarity to the representation of each of these 5 samples and take an average over these similarities as the final score). 

Finally, for the in-context learning specific baselines, when these baselines need a base model for optimising the selection (e.g., evaluating the selected samples as part of the method or running different steps), we use all the same models as during all of the experiments (Mistral, Zephyr, LLaMA, Gemma and Qwen). In other words, we select a set of samples for each of the models as done for the gradient-based few-shot learning as well. This also includes running a hyperparameter search informed by the recommendations of the authors to identify the optimal setup for the search. In addition, whenever necessary, we modify the source code to allow for running the method.

For the core-set selection strategies, we use the modified implementation of the methods released by~\citet{guo2022deepcore} and its extension provided by~\citet{park2022active} for active learning methods. For the Cartography strategy, we use the modified implementation released by the original authors~\citet{swayamdipta-etal-2020-dataset}.

\section{Experiments Setup: Further Details}
\label{app:experimental-setup-details}

The experiments and evaluation of different sample selection strategies are done on the within-domain few-shot learning, where the classes in training, validation and testing datasets all come from the same domain (e.g., from the same dataset). Each evaluation run is done on 600 randomly selected tasks for the gradient few-shot learning approaches and 300 tasks for the in-context learning approaches (due to high computation costs of LLMs). Each task is a 5-way 5-shot (or $K$-shot for experiments where we change the number of shots) classification task evaluated on 16 test samples per class, following the standard few-shot learning methodology and the evaluation from~\citet{meta-album-2022}. The performance of a single evaluation run is reported as the mean accuracy across these tasks and the different splits of data into labelled/unlabelled and train-validation-test sets (we use 5 splits of data, where in each we split the available data into train-validation-test sets and select 200 samples per class for training, as mentioned in the main part of the paper). These scores are then used to report the findings in the main experiments, either as mean and standard deviation over multiple repeated runs of the specific sample selection strategy or over the results from different models and datasets to produce the aggregated figures. As such, we do not explicitly report the standard deviation caused by the different data splits or different sampled tasks. However, the observed deviation from these two factors is quite small (on average up to $0.1$ in terms of accuracy), as it is calculated over large number of repeats. This is also the reason why we can observe zero standard deviation on some of the sample selection strategies.

\begin{table*}[!tbh]
\centering
\small
\caption{Prompt formats and verbalisers used for different datasets in the paper. The News classification datasets include \textbf{20 News Group} and \textbf{News Category} datasets and Intent classification includes \textbf{ATIS}, \textbf{Facebook}, \textbf{HWU-64} and \textbf{SNIPS}. The \textit{[Option 1-5]} are replaced with the names of the classes as defined by the verbaliser. The \textit{[Input]} is replaced by the sentence of the samples and the \textit{[Output]} is replaced with the name of class as defined by the verbaliser. The \textit{[Input]} and \textit{[Output]} are repeated for each in-context sample, while the final \textit{[Output]} is used to determine the predicted class. The same format is used for all in-context learning models as it was the best performing one from the ones tried during basic prompt engineering and following the recommendations from~\cite{sun2023pushing} and our own experience.}
\label{tab:prompt-format}
\begin{tabularx}{\textwidth}{@{}lX@{}}
\toprule
\textbf{Datasets} & \textbf{Prompt Format} \\ \midrule
News classification   & \begin{tabular}[c]{@{}l@{}}Determine category of the sentence using following options:\\1) \textit{[Option 1]} 2) \textit{[Option 2]} 3) \textit{[Option 3]} 4) \textit{[Option 4]} 5) \textit{[Option 5]}.\\ \textit{[Input]} \\ \textit{[Output]} \end{tabular} \\ \midrule

Intent classification   & \begin{tabular}[c]{@{}l@{}}Determine intent of the sentence using following options:\\1) \textit{[Option 1]} 2) \textit{[Option 2]} 3) \textit{[Option 3]} 4) \textit{[Option 4]} 5) \textit{[Option 5]}.\\ \textit{[Input]} \\ \textit{[Output]} \end{tabular} \\ \midrule

\textbf{Dataset} & \textbf{Verbaliser} \\ \midrule

20 News Group &  \{IBM, Middle East Politics, Windows XP, Motorcycles, Medicine, For Sale, Religion, MS Windows, Baseball, Auto, Hockey, Mac, Graphics, Christianity, Guns, Electronics, Space, Crypto, Atheism, Politics\}  \\ 
News Category & \{Politics, World News, Parenting, Money, Wellness, Business, Weddings, Entertainment, Impact, Black Voices, Queer Voices, Crime, Divorce, Food and Drink, Worldpost, Parents, Travel, The Worldpost, Healthy Living, Taste, Media, Culture and Arts, Style, Weird News, Style and Beauty, Comedy, Home and Living, Sports, Environment, Education, Good News, Fifty, Women, Science, Arts and Culture, US News, Green, Tech, Religion, Latino Voices, College, Arts\} \\ 
ATIS & \{Flight, Abbreviation, City, Airfare, Ground Service, Ground Fare, Airline, Flight Number, Aircraft, Distance, Capacity, Flight Time, Quantity, Airport\} \\ 
Facebook & \{Road Condition, Departure, Duration, Event, Arrival, Directions, Distance, Traffic, Update, Combine, Route, Location\} \\ 
HWU-64 & \{Set Alarm, Definition, Play Music, Set Calendar, Play Radio, Confirm, Quirky, Send Email, Currency, Like, Social Query, News, Neutral, Stop, Calendar Query, Alarm Query, Recommend Movie, Play Audiobook, Weather Query, Praise, Social Post, Play Podcast, Negate, Recipe, Affirm, Remove Calendar, Recommend Location, Remove List, Explain, Email Query, Maths, Stock, Transport Query, Order Takeaway, Datetime Query, Repeat, Factoid, Taxi, Add List, Add Contact, Recommend Event, Mute Volume, List Query,  Ticket, Convert Datetime, Joke, Remove Alarm, Traffic, Volume Up, Takeaway Query, Play Game, Contact Query, Music Query, Volume Other, Volume Down, Music Setting, Greet, Dislike\} \\ 
SNIPS & \{Playlist, Weather, Event, Musing, Creative Work, Rate Book, Book Restaurant\} \\ \bottomrule
\end{tabularx}
\end{table*}

For the in-context learning approaches, we use the instruction-tuned Mistral-7B~\citep{jiang2023mistral} (instruct v0.1), Zephyr-7B~\citep{tunstall2023zephyr} (alpha version), LLaMA-3.1-8B~\citep{grattafiori2024llama}, Gemma-3-4B~\citep{gemma_2025} and Qwen-2.5-7B~\citep{qwen2.5} large language models in full precision (although 4-bit precision has shown minimal impact on evaluation). We perform no further training for these models (except in case of sample selection, which is covered in Appendix~\ref{app:sss-details}). Instead we run a basic prompt engineering to find the best performing template for each model. In the end, we opted for the single template that performed the best on average for each model and each dataset, which is illustrated in Table~\ref{tab:prompt-format}, along with the verbaliser (i.e., the possible outputs of the model that are mapped to the classes in the dataset). Finally, the in-context learning models are set to provide deterministic outputs (controlling the temperature and disabling sampling) and set to generate maximum of 10 tokens. In case where multiple words that can be mapped to a dataset class are generated, the output is treated as incorrect as we consider the model to be just hallucinating and not working as intended (this behaviour was mainly observed when using the models in zero-shot or 50-shot setting, where the model either does not understand the task or the context is not sufficient for it to follow the instructions and often either repeats all the classes or generates further sentences).

For the gradient few-shot learning approaches we use a pretrained BERT-base~\citep{devlin-etal-2019-bert} model for the text data and the ResNet-18~\citep{he2016deep} for image data, pretrained on Imagenet dataset~\citep{deng2009imagenet}, as base models. All the models are further trained on each dataset using the different few-shot learning approaches, namely the Prototypical Networks~\citep{snell2017prototypical}, Model-Agnostic Meta-Learning (MAML)~\citep{finn2017model} and Few-Shot Fine-Tuning~\citep{chen2019closer}. In case of Few-Shot Fine-Tuning, the further training is done on the concatenation of all the meta-training classes and their corresponding samples, while at the meta-test time, only the last layer is replaced and fine-tuned on the specific task with 5 classes. The Prototypical Networks and MAML follow the typical few-shot training setting, training on large number of 5-way few-shot tasks. The meta-learning approaches on image data are trained on 15 000 tasks and the Few-Shot Fine-Tuning approach is trained on 15 000 randomly sampled batches of size 16. On text data, the number of sampled tasks and batches is increased to 25 000. In addition, the MAML approach uses 5 training iteration and 10 testing iterations on image data and 10 training and 15 testing on text data, while Prototypical Networks use only 1 in both cases. Finally, all approaches use the Adam optimiser with 0.001 (meta) learning rate and 0.01 base learning rate. The full training process is repeated 3 times and the best performing model, selected using a separate validation dataset, is selected for use throughout the reported experiments. For the LASSO model we use the alpha value of 0.001.

All the hyperparameters for all models were determined using a hyperparameter search on each data modality and model using a single validation split. For the experiments, we use a modified implementation provided by the MetaAlbum~\citep{meta-album-2022}, extended to further approaches, models and datasets. In addition, we also use the hyperparameters used in the paper as a starting point for the hyperparameter search.

\section{Additional Results: Table Results from Text Datasets}
\label{app:table-text}

In this appendix, we provide the results from the main experiments in table form -- in Table~\ref{tab:strategies-diff-text}.

\begin{table*}[tbh]
\centering
\small
\caption{Benefit of the different selection strategies calculated as the difference in accuracy to the \textit{classic} few-shot selection strategy, aggregated over the text datasets. The subscript represents the standard deviation of the difference over the aggregated datasets.}
\label{tab:strategies-diff-text}
{\resizebox{\textwidth}{!}{
\begin{tabular}{@{}lcccccccc@{}}
\toprule
\textsc{Strategy}          & \multicolumn{5}{c}{\textsc{Text Data - ICL}} & \multicolumn{3}{c}{\textsc{Text Data - FSL}}       \\
       & \textsc{Mistral} & \textsc{Zephyr} & \textsc{LLaMA} & \textsc{Gemma} & \textsc{Qwen}                   & \textsc{ProtoNet} & \textsc{MAML} & \textsc{FT}                \\ \midrule
Classic & $+0.00_{0.00}$        & $+0.00_{0.00}$        & $+0.00_{0.00}$        & $+0.00_{0.00}$        & $+0.00_{0.00}$ & $+0.00_{0.00}$        & $+0.00_{0.00}$        & $+0.00_{0.00}$        \\
Random  & $+0.46_{0.86}$        & $+0.01_{0.82}$        & $+0.25_{1.10}$        & $-0.04_{1.07}$        & $+0.38_{0.89}$ & $+0.10_{0.30}$        & $+0.07_{0.21}$        & $-0.26_{0.11}$         \\ \midrule
Similarity      & $-0.24_{1.11}$        & $-1.22_{1.61}$        & $-0.77_{0.81}$        & $-0.41_{0.76}$        & $+0.08_{0.92}$ & $-2.15_{1.82}$        & $-3.51_{1.50}$        & $-1.56_{1.45}$        \\
Diversity       & $-0.66_{1.08}$        & $-0.42_{0.80}$        & $-0.23_{1.19}$        & $-0.63_{1.35}$        & $+0.03_{0.95}$ & $-1.65_{1.02}$        & $-2.77_{1.55}$        & $-3.65_{2.69}$        \\ \midrule
Entropy & $+1.02_{0.98}$        & $+0.70_{0.38}$        & $+1.32_{0.91}$        & $+1.25_{1.12}$        & $+1.02_{0.69}$ & $-0.17_{0.33}$        & $-0.20_{0.68}$        & $+0.07_{1.03}$        \\
Margin  & $+0.93_{0.94}$        & $+0.51_{0.50}$        & $+1.59_{0.94}$        & $+1.34_{0.97}$        & $+1.02_{0.61}$ & $+0.12_{0.10}$        & $+0.11_{0.17}$        & $+0.18_{0.35}$        \\
Least Confidence        & $+0.49_{1.03}$        & $+0.04_{1.17}$        & $+0.38_{1.27}$        & $+0.00_{1.15}$        & $+0.57_{1.10}$ & $+0.02_{0.25}$        & $-0.25_{0.72}$        & $-0.14_{0.59}$        \\
Loss    & $-0.56_{0.88}$        & $-0.65_{0.83}$        & $-0.26_{1.34}$        & $-0.59_{1.15}$        & $-0.03_{1.17}$ & $-1.49_{0.89}$        & $-3.00_{1.46}$        & $-3.75_{1.92}$        \\ \midrule
CAL     & $+0.39_{1.27}$        & $+0.18_{1.56}$        & $+0.58_{1.17}$        & $-0.04_{1.48}$        & $+0.35_{1.23}$ & $-0.18_{0.18}$        & $-0.76_{0.69}$        & $-0.69_{0.54}$        \\
Craig   & $+0.66_{1.36}$        & $-0.38_{1.03}$        & $+1.03_{1.10}$        & $+0.33_{0.98}$        & $+0.47_{1.22}$ & $+0.22_{0.40}$        & $-0.15_{0.69}$        & $-0.00_{0.62}$        \\
DeepFool       & $+0.40_{1.29}$        & $+0.18_{1.52}$        & $+0.88_{1.25}$        & $+0.25_{1.59}$        & $+0.44_{1.18}$  & $-0.18_{0.18}$        & $-0.90_{0.65}$        & $-0.76_{0.59}$        \\
Forgetting      & $+1.27_{0.96}$        & $+0.77_{0.41}$        & $+1.46_{0.88}$        & $+0.81_{0.95}$        & $+1.17_{0.70}$ & $+0.39_{0.26}$        & $+0.49_{0.54}$        & $+0.41_{0.42}$        \\
Glister & $+0.64_{0.98}$        & $+0.21_{0.92}$        & $+0.60_{1.41}$        & $+0.29_{1.60}$        & $+0.48_{1.19}$ & $+0.19_{0.43}$        & $+0.06_{0.49}$        & $+0.03_{0.47}$        \\
Grand   & $+0.29_{0.62}$        & $-0.54_{1.41}$        & $+0.70_{1.60}$        & $+0.29_{1.43}$        & $+0.14_{0.87}$ & $-0.01_{0.38}$        & $-0.27_{0.34}$        & $-0.40_{0.67}$        \\
Herding & $+0.24_{0.91}$        & $+0.11_{0.78}$        & $-0.26_{1.26}$        & $-0.00_{1.32}$        & $+0.15_{1.11}$ & $+0.03_{0.51}$        & $-0.24_{0.69}$        & $-0.04_{0.45}$        \\
KCenter & $+0.13_{1.24}$        & $+0.31_{0.88}$        & $-0.30_{1.39}$        & $+0.41_{1.18}$        & $+0.23_{1.02}$ & $+0.04_{0.58}$        & $-0.15_{0.91}$        & $+0.03_{0.49}$        \\
Graph-Cut       & $+0.42_{1.11}$        & $-0.17_{0.92}$        & $+0.59_{1.15}$        & $+0.25_{1.00}$        & $+0.83_{0.99}$ & $+0.34_{0.50}$        & $+0.35_{0.44}$        & $+0.10_{0.33}$        \\ \midrule
Cartography\textsubscript{Easy} & $+0.16_{0.76}$        & $+0.06_{0.98}$        & $+0.60_{0.85}$        & $+0.04_{0.96}$        & $+0.23_{1.08}$ & $-0.04_{0.17}$        & $+0.02_{0.21}$        & $-0.19_{0.31}$        \\
Cartography\textsubscript{Ambiguous}    & $+0.21_{0.64}$        & $+0.02_{0.64}$        & $+0.04_{0.61}$        & $-0.26_{0.68}$        & $+0.40_{0.26}$ & $-0.06_{0.12}$        & $-0.01_{0.13}$        & $-0.06_{0.21}$        \\
Cartography\textsubscript{Hard} & $+1.60_{0.93}$        & $+1.26_{0.39}$        & $+1.82_{1.18}$        & $+1.53_{1.38}$        & $+1.47_{0.70}$ & $+0.05_{0.37}$        & $+0.08_{0.29}$        & $-0.14_{0.28}$        \\
Cartography\textsubscript{Easy+Ambig.}  & $+0.21_{0.64}$        & $+0.02_{0.64}$        & $-0.33_{0.93}$        & $+0.15_{0.78}$        & $+0.02_{0.59}$ & $+0.59_{0.30}$        & $+0.78_{0.39}$        & $+0.37_{0.16}$        \\ \midrule
LENS   & $+1.73_{0.70}$        & $+1.42_{0.32}$        & $+1.57_{1.31}$        & $+1.61_{1.41}$        & $+0.87_{0.91}$  & N/A   & N/A   & N/A   \\
Active Prompt  & $+1.89_{0.71}$        & $+1.59_{0.36}$        & $+1.72_{1.32}$        & $+1.89_{1.28}$        & $+1.05_{0.85}$  & N/A   & N/A   & N/A   \\
EXPLORA & $+2.17_{1.00}$        & $+1.81_{0.38}$        & $+1.98_{1.29}$        & $+2.23_{1.10}$        & $+1.60_{0.68}$ & N/A   & N/A   & N/A   \\
CASE    & $+2.40_{1.07}$        & $+2.10_{0.56}$        & $+2.33_{1.28}$        & $+2.47_{1.11}$        & $+1.87_{0.71}$ & N/A   & N/A   & N/A   \\ \midrule
ACSESS\textsubscript{Uniform}   & $+2.30_{1.11}$       & $+1.93_{0.48}$        & $+2.08_{1.30}$        & $+2.21_{1.08}$        & $+1.78_{0.70}$ & $+0.92_{0.41}$        & $+1.34_{0.45}$        & $\boldsymbol{+1.36_{0.75}}$        \\
ACSESS\textsubscript{Weighted}  & $\boldsymbol{+2.55_{1.08}}$       & $\boldsymbol{+2.15_{0.53}}$        & $\boldsymbol{+2.48_{1.25}}$        & $\boldsymbol{+2.73_{1.13}}$        & $\boldsymbol{+2.09_{0.74}}$ & $\boldsymbol{+1.02_{0.44}}$        & $\boldsymbol{+1.49_{0.47}}$        & $+1.07_{0.69}$        \\
ACSESS\textsubscript{With Random}       & $+1.78_{0.98}$        & $+1.42_{0.45}$        & $+1.85_{1.38}$        & $+1.55_{1.05}$        & $+0.99_{0.73}$ & $+0.72_{0.33}$        & $+1.03_{0.48}$        & $+0.80_{0.54}$        \\

\bottomrule
\end{tabular}
}}
\end{table*}

\section{Additional Experiments: Performance on Image Data}
\label{app:image-results}

In this appendix, we present the setup and results of the main experiments done on image data. The datasets are chosen similarly to text datasets, where we focus on datasets with different characteristics, such as class imbalance and noisy datasets. Specifically, we use 8 datasets from the \textbf{Meta-Album}~\citep{meta-album-2022}, specifically \textbf{LR\_AM.DOG} and \textbf{LR\_AM.AWA} for animal classification, \textbf{VCL.APL} for plane classification, \textbf{HUM\_ACT.ACT\_410} for classification of human actions, \textbf{MNF.TEX\_DTD} for classification of texture, \textbf{MCR.PRT} and \textbf{MCR.PNU} for classification of microscope images, and \textbf{PLT.FLW} for flower classification.

\begin{figure*}[t!]
    \centering
    \includegraphics[width=1\linewidth]{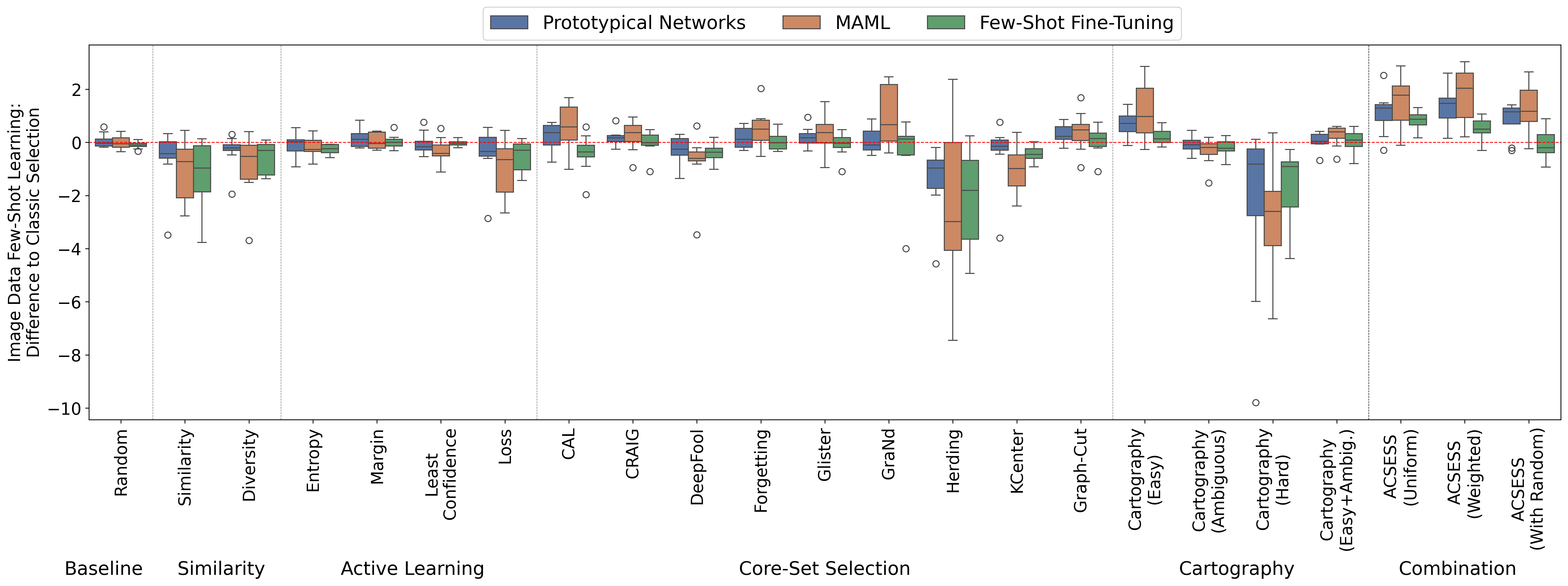}
    \caption{Benefit of the different sample selection strategies, calculated as the difference in accuracy between the specific strategy and the classic few-shot selection, aggregated over the image datasets (box plots show the distribution of results across the datasets). The performance of the classic selection is represented as the red dashed line (zero value). The consistently beneficial selection strategies depend on the approach (meta-learning vs. fine-tuning). Our proposed method ACSESS consistently leads to improved performance.}
    \label{fig:strategies-diff-image}
\end{figure*}

The results of the main experiments, where we investigate the impact of different single-property sample selection strategies on few-shot learning, as well as provide the comparison with the proposed ACSESS method, are presented in an aggregated form in Table~\ref{tab:strategies-diff-image} and Figure~\ref{fig:strategies-diff-image}. Similarly to experiments on text data, the results for each individual dataset and model are presented in supplementary material.

\begin{table*}[tbh]
\centering
\small
\caption{Benefit of the different selection strategies calculated as the difference in accuracy to the \textit{classic} few-shot selection strategy, aggregated over the image datasets. The subscript represents the standard deviation of the difference over the aggregated datasets.}
\label{tab:strategies-diff-image}
\tabcolsep=0.75cm
\begin{tabular}{@{}lccc@{}}
\toprule
\textsc{Strategy}        & \multicolumn{3}{c}{\textsc{Image Data - FSL}}      \\
        & \textsc{ProtoNet} & \textsc{MAML} & \textsc{FT}              \\ \midrule
Random  & $+0.06_{0.26}$        & $-0.00_{0.27}$        & $-0.10_{0.13}$\\
Similarity      & $-0.64_{1.14}$        & $-1.05_{1.15}$        & $-1.21_{1.24}$ \\
Diversity       & $-0.35_{0.65}$        & $-0.93_{1.22}$        & $-0.57_{0.58}$ \\ \midrule
Entropy & $-0.12_{0.43}$        & $-0.17_{0.37}$        & $-0.26_{0.18}$        \\
Margin  & $+0.15_{0.34}$        & $+0.05_{0.29}$        & $+0.02_{0.26}$        \\
Least Confidence        & $-0.05_{0.41}$        & $-0.31_{0.46}$        & $-0.03_{0.12}$  \\
Loss    & $-0.44_{1.01}$        & $-1.02_{1.07}$        & $-0.53_{0.57}$        \\  \midrule
CAL     & $+0.19_{0.54}$        & $+0.54_{0.88}$        & $-0.43_{0.72}$        \\
CRAIG   & $+0.18_{0.29}$        & $+0.26_{0.60}$        & $-0.04_{0.45}$        \\
DeepFool        & $-0.29_{0.51}$        & $-0.77_{1.11}$        & $-0.41_{0.36}$\\
Forgetting      & $+0.16_{0.37}$        & $+0.53_{0.72}$        & $+0.03_{0.34}$\\
Glister & $+0.20_{0.37}$        & $+0.33_{0.73}$        & $-0.10_{0.45}$         \\
GraNd   & $+0.09_{0.49}$        & $+0.95_{1.11}$        & $-0.42_{1.41}$         \\
Herding & $-1.43_{1.31}$        & $-2.36_{2.94}$        & $-2.14_{1.78}$        \\
KCenter & $-0.45_{1.24}$        & $-1.03_{0.87}$        & $-0.42_{0.31}$        \\
Graph-Cut      & $+0.32_{0.36}$        & $+0.40_{0.75}$        & $+0.05_{0.53}$ \\  \midrule
Cartography\textsubscript{Easy}    & $+0.65_{0.50}$        & $+1.17_{1.11}$        & $+0.21_{0.28}$  \\
Cartography\textsubscript{Ambiguous}       & $-0.08_{0.36}$        & $-0.36_{0.50}$        & $-0.21_{0.32}$  \\
Cartography\textsubscript{Hard}    & $-2.42_{3.35}$        & $-3.03_{2.26}$        & $-1.71_{1.47}$        \\
Cartography\textsubscript{Easy+Ambig.}     & $+0.05_{0.33}$        & $+0.24_{0.40}$        & $+0.04_{0.40}$ \\  \midrule

ACSESS\textsubscript{Uniform}         & $+1.12_{0.80}$        & $+1.59_{0.97}$        & $\boldsymbol{+0.81_{0.35}}$   \\
ACSESS\textsubscript{Weighted}        & $\boldsymbol{+1.31_{0.75}}$   & $\boldsymbol{+1.83_{0.97}}$   & $+0.49_{0.41}$ \\
ACSESS\textsubscript{With Random}     & $+0.85_{0.65}$        & $+1.32_{0.91}$        & $-0.06_{0.56}$        \\
\bottomrule
\end{tabular}
\end{table*}

Although the results on the image modality are slightly different, the main findings hold true. Many selection strategies perform worse than the \textit{Classic selection} (such as hard to learn samples), while only specific ones lead to consistently better performance (e.g., easy to learn samples or those that are forgotten the least). 

A specific new finding that we can draw is that \textbf{the overall benefit of sample selection is higher for noisy datasets and limited for high-quality datasets.} Sample selection leads to a considerable performance increase over the \textit{Classic selection}, on average around $2$ percentage points, for the datasets with many noisy samples (as determined by the \textit{Cartography}) or for datasets with large class imbalance (e.g., difference between lowest and median number of samples per class in few hundred samples), such as HUM\_ACT.ACT\_410, MCR.PRT or MCR.PNU. At the same time, performance increase is quite limited (on average around $0.1$ percentage points) for the datasets with few noisy samples or no extensive class imbalance, such as the VCL.APL, PLT.FLW or NewsCategory.

When it comes to the proposed ACSESS strategy, we still observe the priority on learnability properties. In the case of the image data, the combination of easy to learn samples (\textit{Cartography}) that are located on the decision boundary (\textit{Margin}) provides the most benefit. This represents the combination of the lowest number of strategies we have observed so far, which indicates that the sample selection strategies are well-designed for image data, or that the image datasets contain high-quality samples that are easier to identify.

\begin{figure*}
\centering
\begin{subfigure}{.485\textwidth}
  \centering
  \includegraphics[width=1\linewidth]{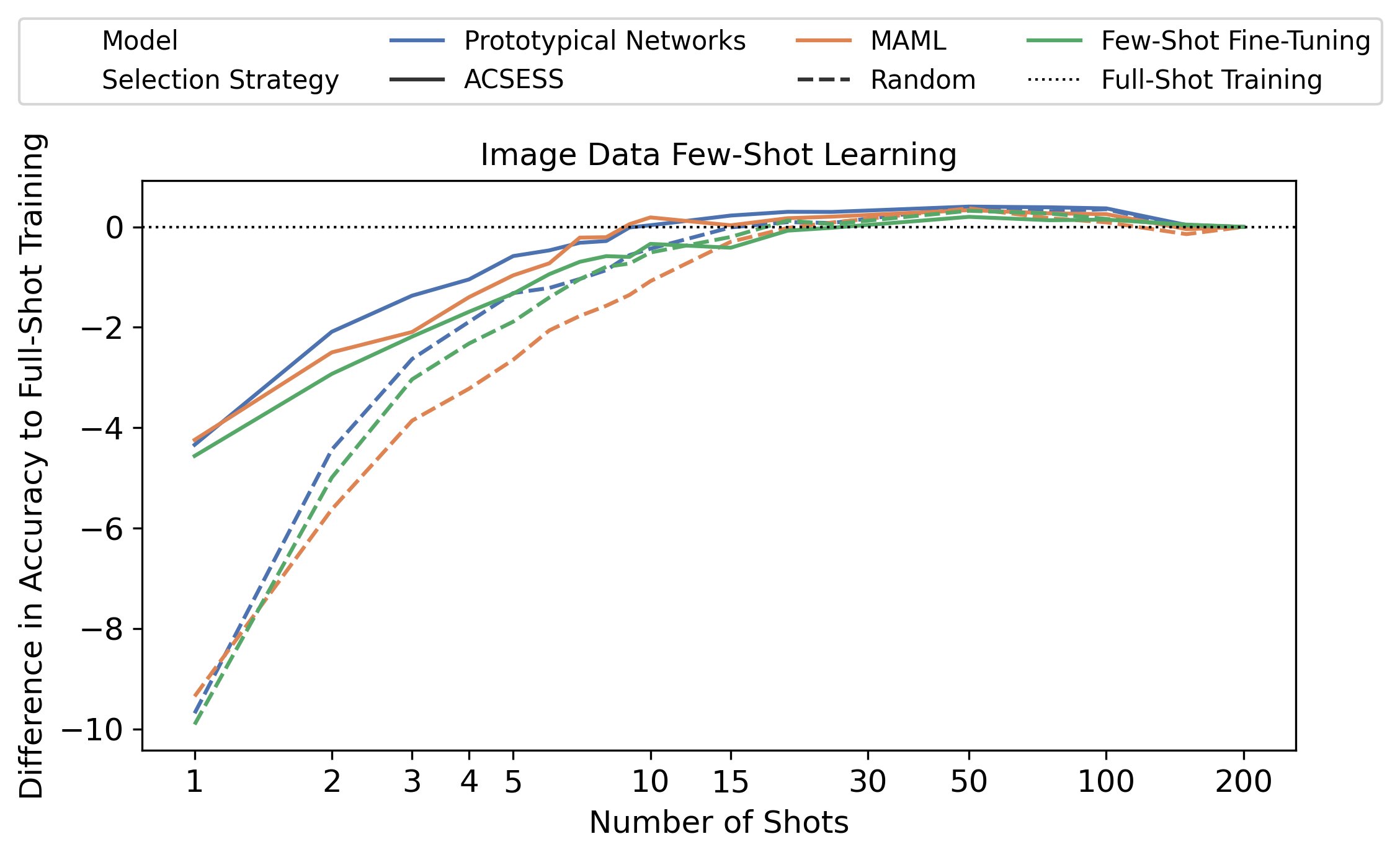}
  \label{subfig:image-shots}
\end{subfigure}%
\begin{subfigure}{.5\textwidth}
  \centering
  \includegraphics[width=1\linewidth]{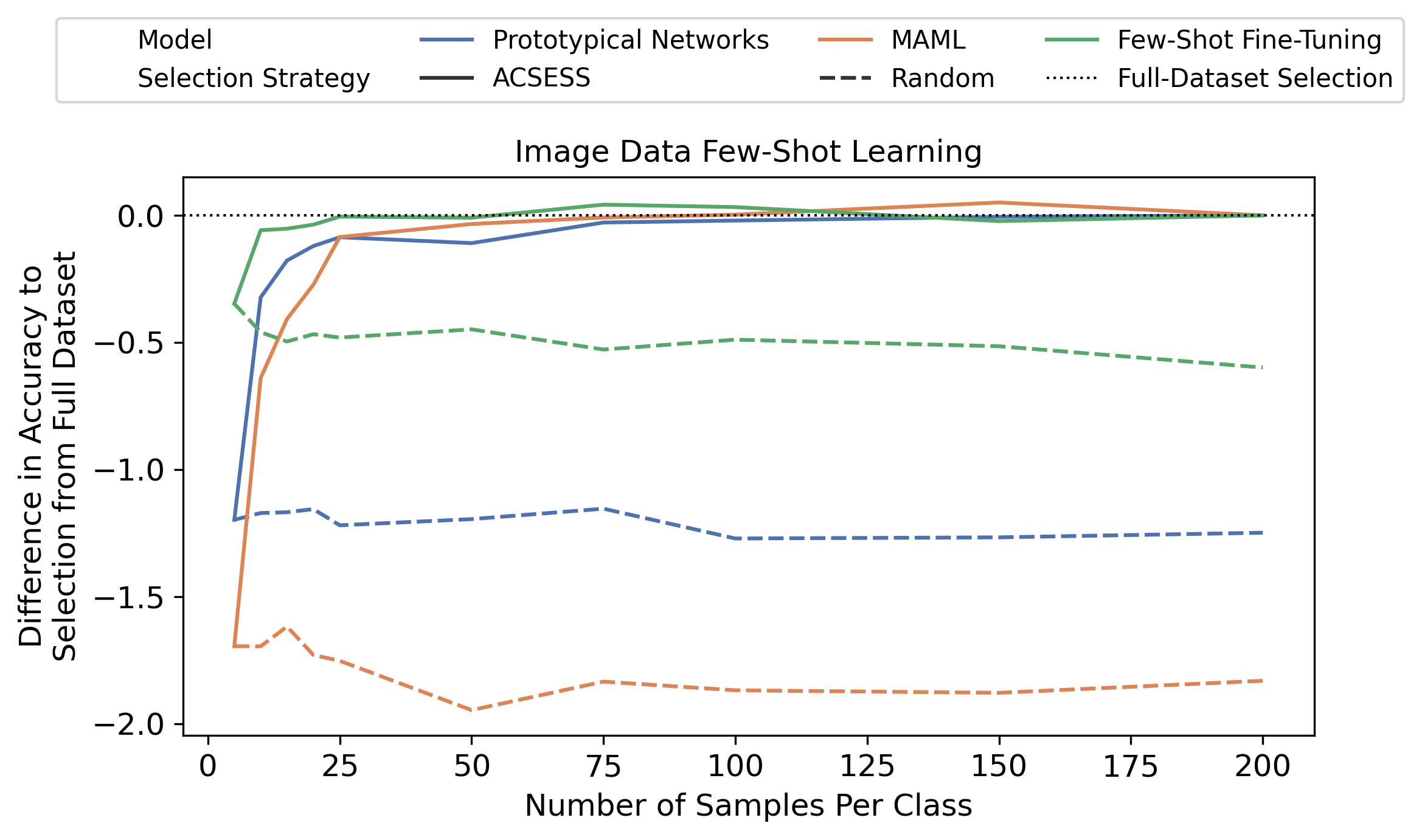}
  \label{subfig:image-datasets}
\end{subfigure}
\caption{The change in benefit of selection strategies based on the number of shots (left) and the difference in accuracy of selection strategies when using dataset subsets of different sizes (right) aggregated over the image datasets.}
\label{fig:image-shots-datasets}
\end{figure*}

Finally, when it comes to the impact of increasing the number of shots, or using datasets of different sizes, we still observe similar findings on image data (see Figure~\ref{fig:image-shots-datasets}). When changing the number of shots, the highest performance is achieved around $10-30$ shots and then remains the same across all of further increases. Similarly, when selecting from datasets of smaller sizes, we can observe the same performance when using as few as $50$ samples (as compared to using the full dataset). At the same time, we observe that the more sophisticated selection strategies outperform random selection most of the time, while degrading to random selection on a high number of shots.

\section{Additional Experiments: Dataset and Model Dependence of Sample Selection Strategies and their Sensitivity to Random Components}
\label{app:additional_insights_dependence_sensitivity}

\begin{figure*}[tbh]
    \centering
    \includegraphics[width=1.0\linewidth]{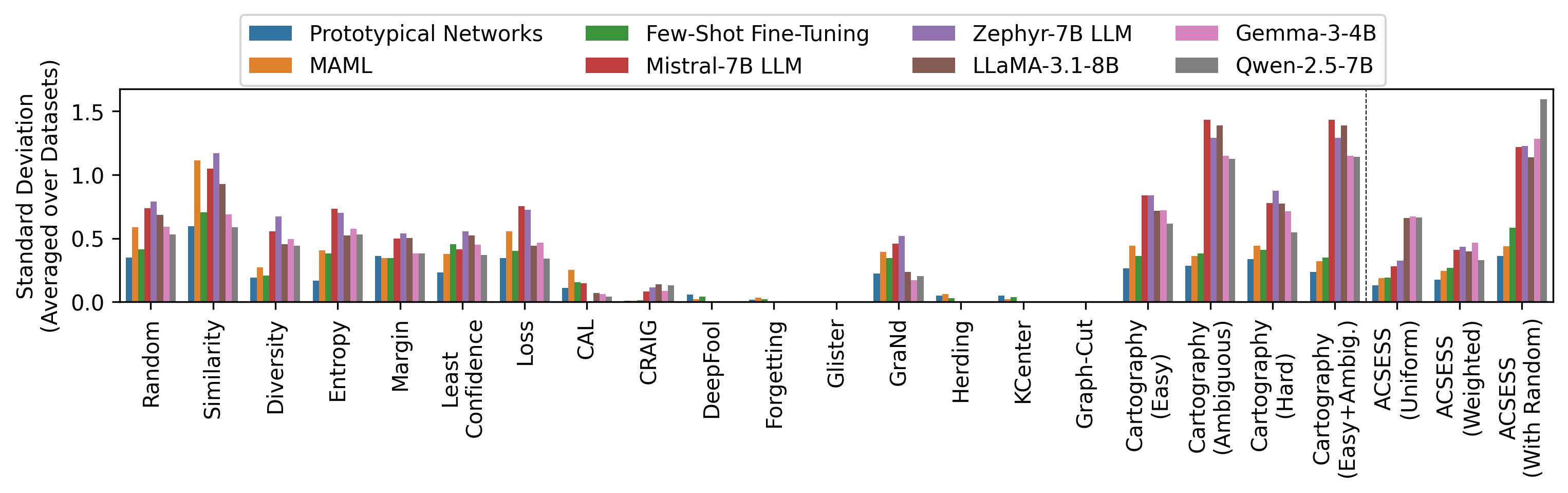}
    \caption{Standard deviation introduced by multiple runs of the different selection strategies. The results for Prototypical Networks, MAML and Few-Shot Fine-Tuning are aggregated over both the image and text datasets, while the results for Mistral and Zephyr are only from text datasets. The ACSESS method shows lower sensitivity to repeated runs compared to the majority of the strategies included in the combination, but still higher sensitivity than specific core set strategies that always lead to the same set of samples.}
    \label{fig:strategy-deviation}
\end{figure*}

In this appendix, we provide additional insights from ablation studies. We are mainly focus on: 1) how the impact of selection strategies is affected by the used models and datasets, i.e., look at the model and dataset dependence of different selection strategies and configurations of our proposed method (based on dataset specific results); and 2) how sensitive the strategies are to their random components, i.e., look at how the set of selected samples changes when the strategies are run multiple times and how this affects the overall performance (based on the variance from repeated strategy runs in Figure~\ref{fig:strategy-deviation}).

\textbf{Specific strategies show large sensitivity to the effects of random seed.} Changing the random seed for specific strategies often leads to different set of selected samples and different performance impact. The core set selection strategies are less affected by this randomness. Specific core set strategies (such as \textit{Glister} or \textit{Graph-Cut}) consistently select the same set of samples and show no variance in results. Other strategies core set strategies (e.g., \textit{Forgetting}, \textit{DeepFool} or \textit{Herding}) show only small variance. On the other hand, strategies with random initial condition, such as the similarity selection or some of the active learning strategies, are more sensitive to the effects of the random seed. \textbf{The ACSESS method shows lower sensitivity to repeated runs} (with the exception of weighting with random element), \textbf{but still may lead to different sets of samples based on the random seed} (as the standard deviation is higher than in core set strategies, i.e., up to $0.5$ for the ACSESS method as opposed to $0$ or $0.1$ for majority of the core set strategies).

\textbf{Many selection strategies show strong dataset dependence.} In many cases, the strategies that perform well on one dataset (such as \textit{DeepFool} on the HUM\_ACT.ACT\_410 dataset) may perform quite poorly on other datasets (such as MCR.PRT or MCR.PNU). Due to this dependence, such strategies cannot be used on any dataset without first evaluating them how well they perform on it. \textbf{Only few strategies perform consistently on the majority of the datasets}, such as \textit{Herding} showing poor performance on all image datasets, or \textit{Cartography} showing good performance across all datasets and models (while the most beneficial configuration of this strategy is dependent on data modality). Finally, we \textbf{do not observe a significant dataset dependence of the ACSESS method}, as it identifies the well-performing strategies and leverages their strengths for the specific dataset. This observation holds through for the weighting schemes as well (i.e., both \textit{uniform} and \textit{weighted} combinations show consistent performance across datasets).

\textbf{Many selection strategies show strong model dependence and sensitivity to selected samples.} The strategies performing well for one model (\textit{GraNd} on MCR.PNU for MAML, or \textit{KCenter} on 20 News for Zephyr) may result in a decrease of performance for other models (\textit{GraNd} for Few-Shot Fine-Tuning, or \textit{KCenter} for Mistral on the same setting). We do not observe such strong dependence for our proposed method ACSESS, with the exception for Few-Shot Fine-Tuning, where the uniform combination is always better than the weighted combination. Overall, having a different set of selected samples, either as a result from using different strategy or from repeated runs of the same strategy, \textbf{affects the Few-Shot Fine-Tuning less}, often leading to smaller benefit. At the same time, the \textbf{MAML} method and the \textbf{in-context learning} models (Mistral and Zephyr) are \textbf{more affected by the selected samples}, leading to more varied results and often higher performance increase. Such sensitivity to sample selection was observed in previous works as well~\citep{pecher2023effects, koksal2022meal, zhang-etal-2022-active, li-qiu-2023-finding}. Curiously, Prototypical Networks gain quite large performance increase from sample selection, but show low sensitivity to the selected samples and variance from repeated runs of sample selection. As such, \textbf{sample selection is more important for MAML and In-Context Learning and less important for Few-Shot Fine-Tuning.}

\section{Additional Experiments: Comparison of Sample Selection Impact Between 5-Shot and 10-Shot Setting}
\label{app:5shot-vs-10shot}

\begin{figure*}[tbh]
    \centering
    \includegraphics[width=1\linewidth]{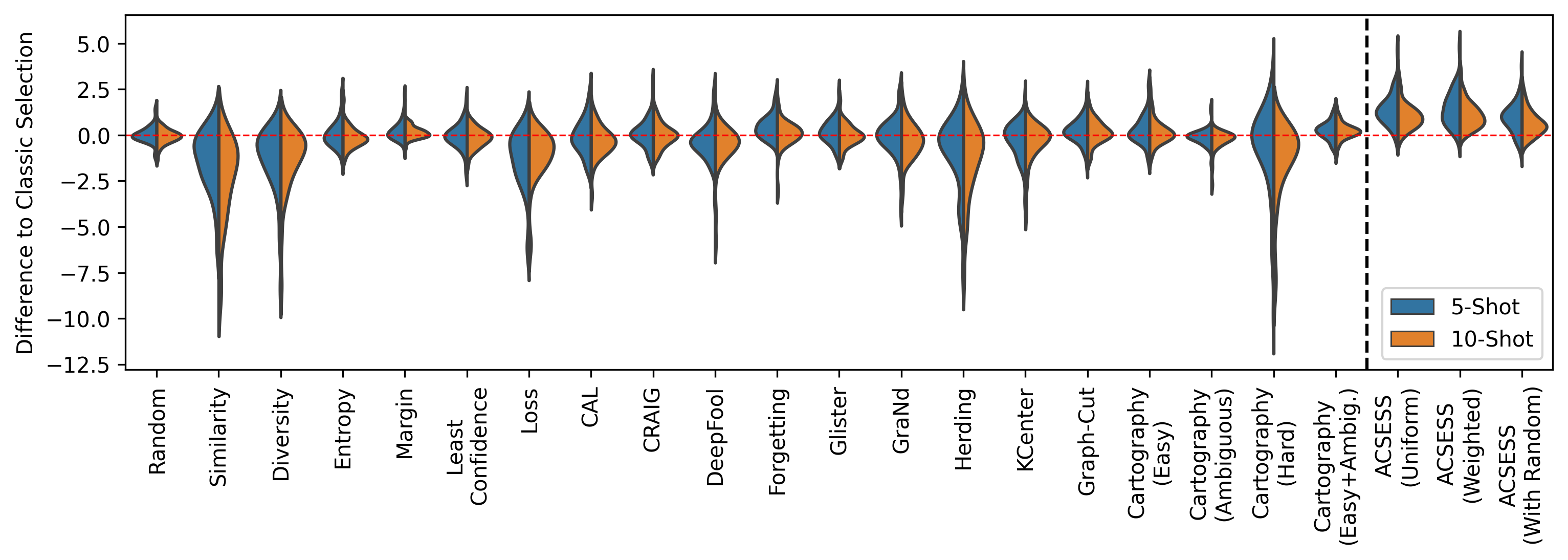}
    \caption{Distribution of the selection strategies benefit (calculated as difference to the classic few-shot selection) for the 5-shot and 10-shot setting. The benefit of the different selection strategies is more significant at lower number of shots.}
    \label{fig:5-shot-vs-10-shot}
\end{figure*}

To better explore the effect of increasing number of selected shots on the sample selection strategies, we repeat the evaluation of all the selection strategies in the 10-shot setting (using the same setup and methodology) and compare the results to 5-shot setting. For both 5-shot and 10-shot setting, we provide a distribution of the results as a difference to the \textit{Classic} selection and illustrate the comparison of these distributions in Figure~\ref{fig:5-shot-vs-10-shot}.

Overall, when moving from 5-shot to 10-shots, we observe an increase in the overall performance of the models for all the selection strategies (with the exception of some bad performing ones) as well as the \textit{Classic} selection. At the same time, the increase in the performance of the \textit{Classic} selection is higher, leading to lower difference and lower impact of the different evaluated selection strategies. This finding holds for all strategies across all models and few-shot learning approaches.

In addition, we also compare the results using Wilcoxon signed-rank test, between the results of 5-shot and 10-shot setting. The resulting p-value ($1.93e-07$) indicates that the different to Classic selection is more significant in case of 5-shot setting.

All in all, these experiments further reinforce the findings from comparison of ACSESS and \textit{Random} selection, where we observe that the the sample selection progressively leads to lower impact as we increase the number of shots we select.

\section{Additional Experiments: Impact of Increasing Number of Shots for Individual Datasets}
\label{app:number_of_shots_deag}

\begin{figure*}[tbh]
    \centering
    \includegraphics[width=.8\linewidth]{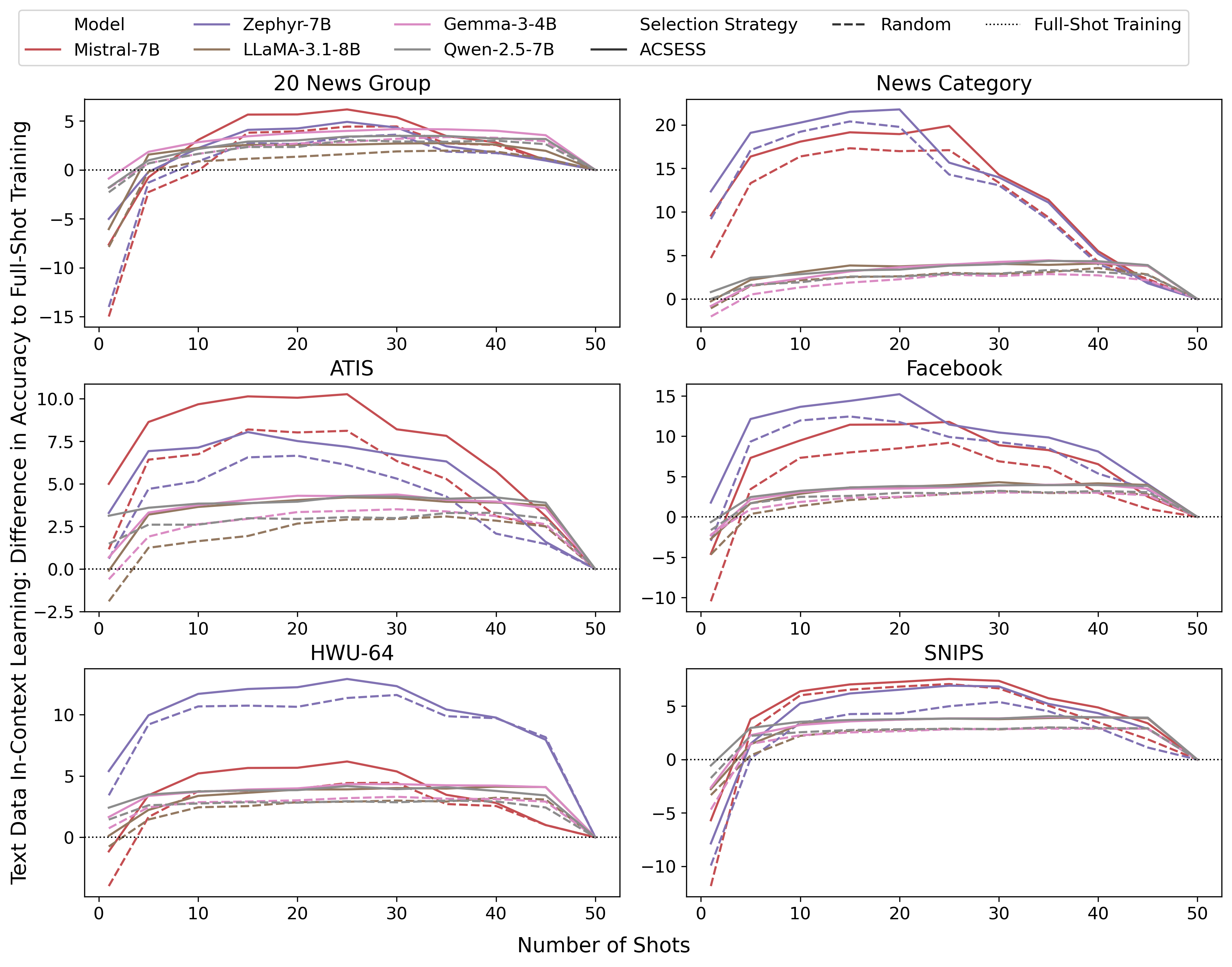}
    \caption{The different benefit of the selection strategies based on the number of selected shots for the text datasets using in-context learning. The reported benefit is calculated as a difference in performance to the one when using the full dataset, for both the ACSESS method and the Random selection.}
    \label{fig:icl-change_by_shots}
\end{figure*}

\begin{figure*}[tbh]
    \centering
    \includegraphics[width=.8\linewidth]{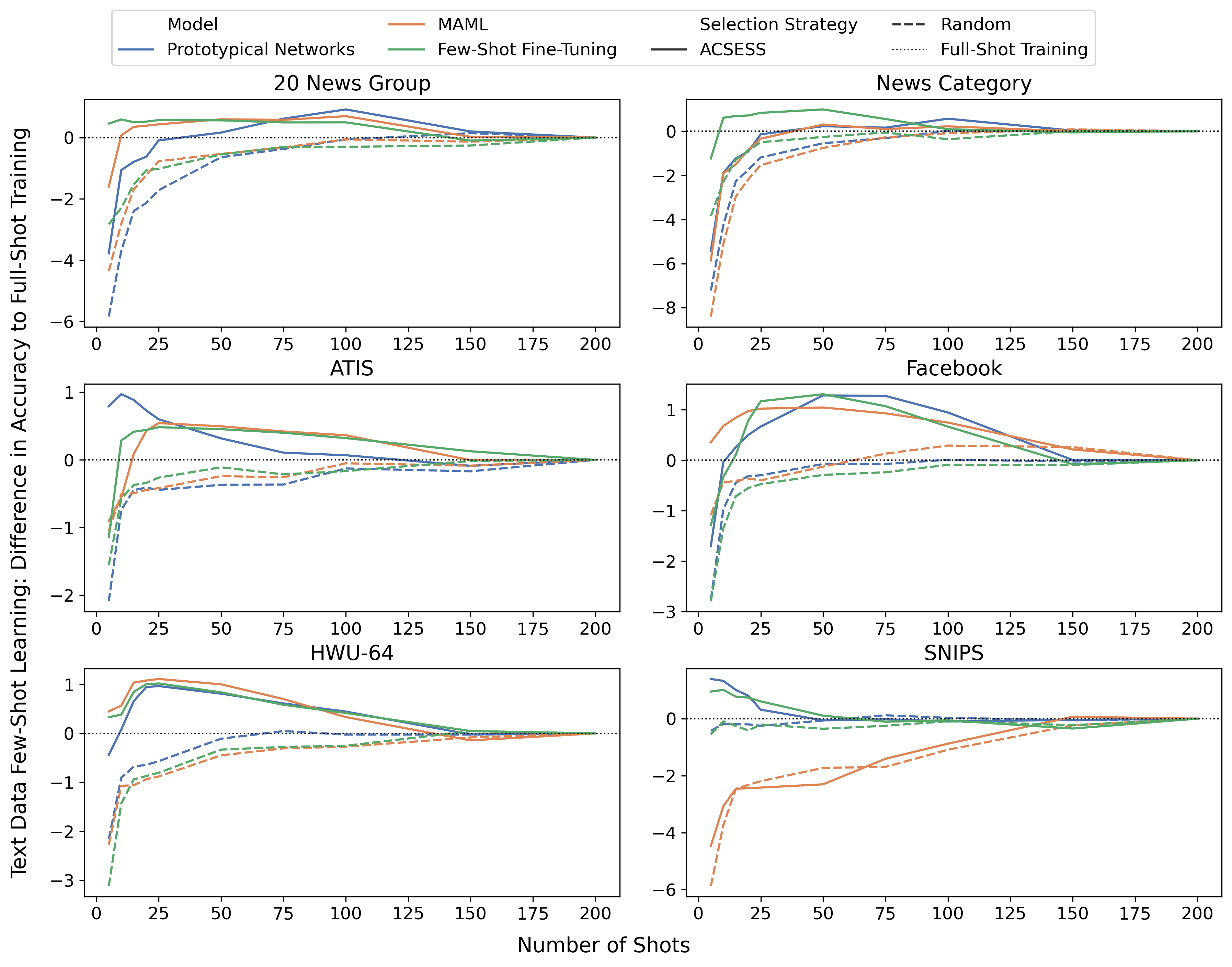}
    \caption{The different benefit of the selection strategies based on the number of selected shots for the text datasets using gradient few-shot learning. The reported benefit is calculated as a difference in performance to the one when using the full dataset, for both the ACSESS method and the Random selection.}
    \label{fig:text-change_by_shots}
\end{figure*}

\begin{figure*}[tbh]
    \centering
    \includegraphics[width=.8\linewidth]{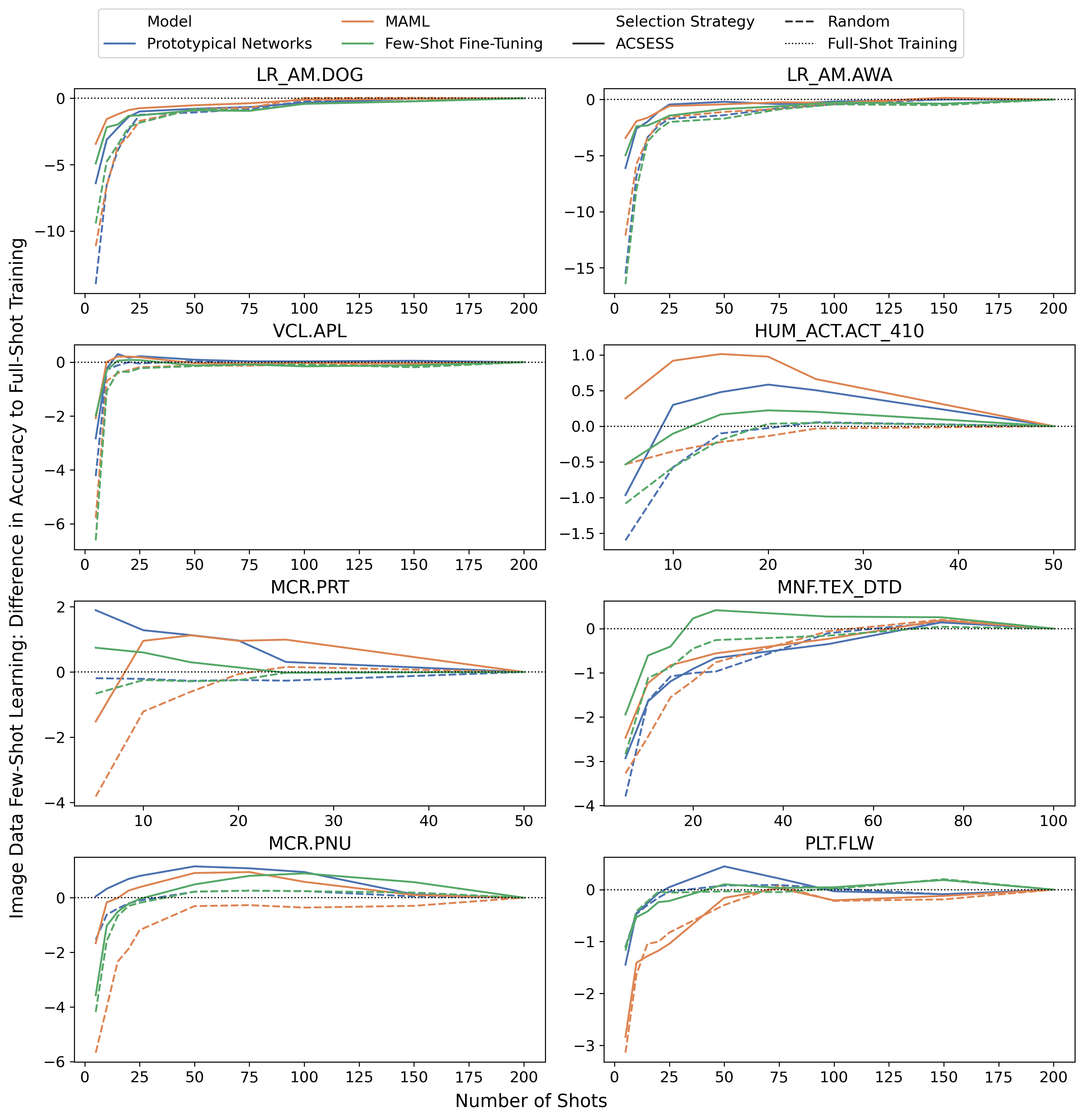}
    \caption{The different benefit of the selection strategies based on the number of selected shots for the image datasets using gradient few-shot learning. The reported benefit is calculated as a difference in performance to the one when using the full dataset, for both the ACSESS method and the Random selection.}
    \label{fig:image-change_by_shots}
\end{figure*}

In this appendix, we provide results from investigating the impact of increasing the number of shots on the benefit of sample selection separately for the individual dataset. For each dataset, we increase the number of selected shots, starting from 1-shot up to the full dataset. The datasets HUM\_ACT.ACT\_410 and MCR.PRT contain only up to 50 samples per class when all samples are used and the MNF.TEX\_DTD only up to 100. All the remaining datasets either contain at maximum 200 samples, or are subsampled to 200 samples per class if necessary (this is especially done for the text datasets, which are often significantly larger). For in-context learning, we increase the number of shots only up to 50 samples per class, as increasing the number further resulted in significantly lower performance (mainly due to limited context size of the large language models). The results for the image datasets are included in Figure~\ref{fig:image-change_by_shots}, for the text datasets using the gradient few-shot learning in Figure~\ref{fig:text-change_by_shots} and for text datasets using the in-context learning in Figure~\ref{fig:icl-change_by_shots}.

For the majority of the datasets, we observe that the highest performance of the sample selection using the ACSESS method is achieved on lower number of shots and then either remains the same until the full dataset, or gradually decreases as we use more samples. Only for specific datasets (LR\_AM.DOG and LR\_AM.AWA datasets for all approaches; PLT.FLW and SNIPS for MAML; and MNF.TEX\_DTD for Prototypical Networks and MAML) we observe continual increase of performance when using sample selection as we increase the number of shots, with the highest performance being achieved only at the full dataset (or slightly beforehand). In addition, on specific datasets, not all models behave the same, for example, on the MNF.TEX\_DTD dataset, only the Few-Shot Fine-Tuning approach achieves highest performance on lower number of shots (which then monotonically decrease as we increase the number of used samples), while the MAML and Prototypical Networks meta-learning approaches achieve the highest performance only when using almost the full dataset. 

On the other hand, using the Random selection, we often observe monotonic increase in performance as the number of the selected shots increases, with the highest performance being achieved on the full dataset. This behaviour can be observed across all datasets and few-shot learning approaches (with exception of in-context learning). As such, this further reinforces our findings and main motivation that the data-centric approach can be beneficial for the few-shot learning. Curating a subset of the most informative and high-quality samples can often lead to performance on par or better than the one when using all the samples for training.

Finally, we also observe that the benefit of sample selection is more significant at lower number of shots. The difference in performance between the ACSESS method and the Random selection on the lower number of shots is significantly higher and slowly decreases as we use more shots for training. The exception is MAML on the PLT.FLW and SNIPS datasets, or Few-Shot Fine-Tuning on PLT.FLW and MCR.PNU datasets, where the performance of the sample selection is similar to the one using random selection even when using small number of shots. After a certain point, the sample selection using the ACSESS method degrades to Random selection. As such, the focus on selecting high quality samples is paramount when selecting only few labelled samples per class, while random selection can be safely used when using higher number of shots.

\section{Additional Experiments: Selection From Datasets of Different Sizes}
\label{app:dataset_size_change_deag}

\begin{figure*}[tbh]
    \centering
    \includegraphics[width=.8\linewidth]{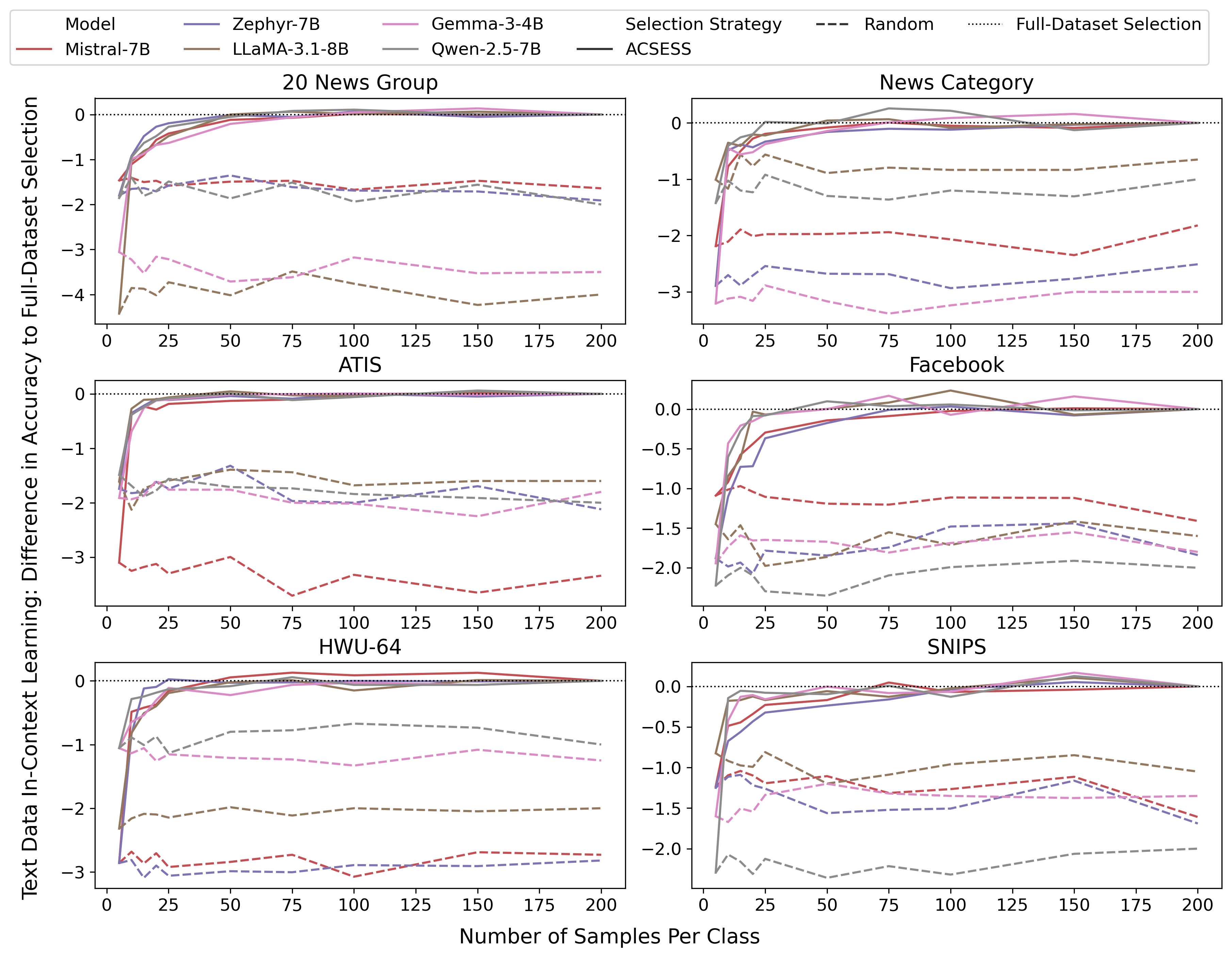}
    \caption{The difference in performance of sample selection strategies on dataset subsets of different sizes for the text datasets using in-context learning. The reported difference is in comparison to using sample selection from the full dataset.}
    \label{fig:icl-dataset_size_change}
\end{figure*}

\begin{figure*}[tbh]
    \centering
    \includegraphics[width=.8\linewidth]{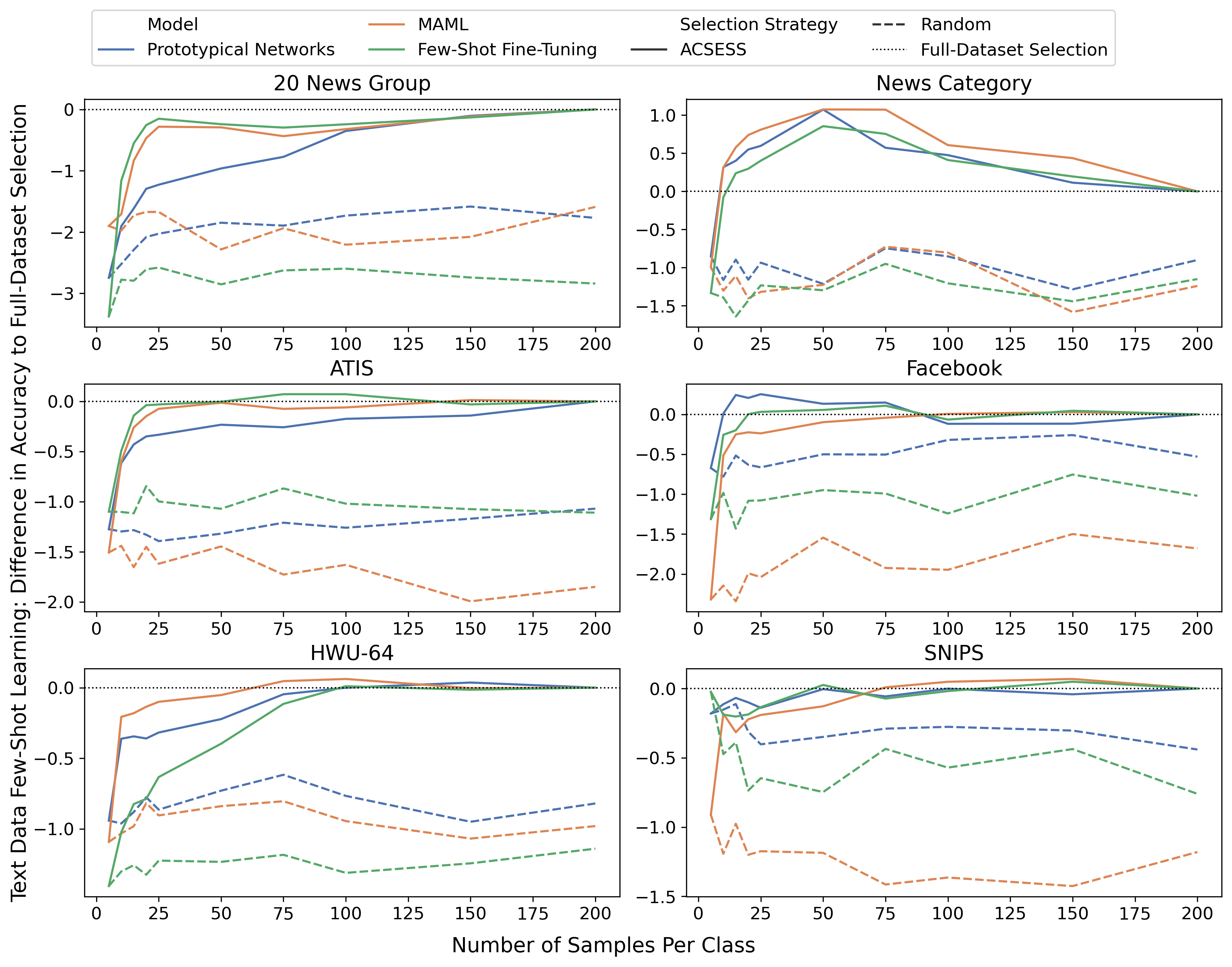}
    \caption{The difference in performance of sample selection strategies on dataset subsets of different sizes for the text datasets using gradient few-shot learning. The reported difference is in comparison to using sample selection from the full dataset.}
    \label{fig:text-dataset_size_change}
\end{figure*}

\begin{figure*}[tbh]
    \centering
    \includegraphics[width=.8\linewidth]{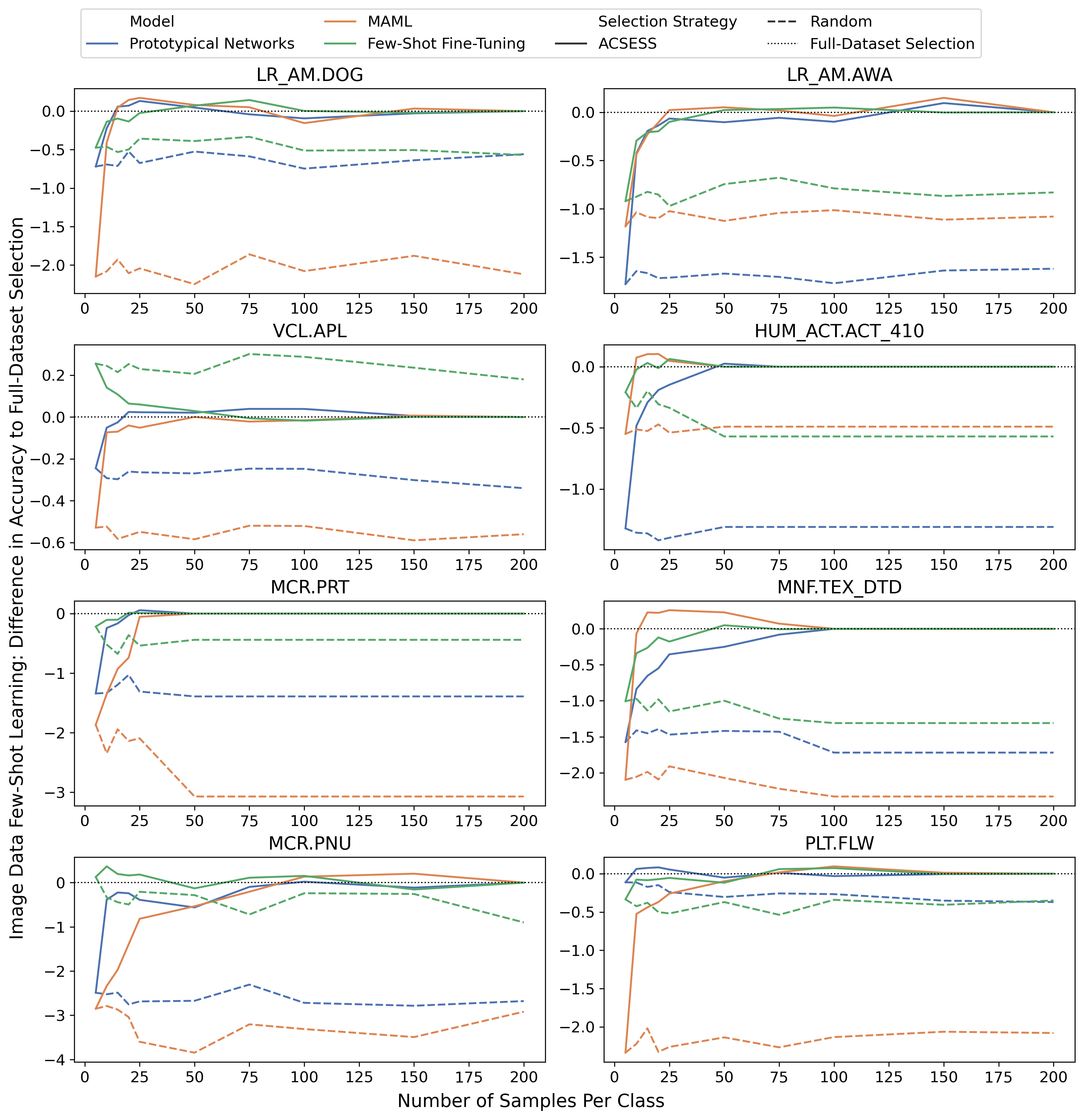}
    \caption{The difference in performance of sample selection strategies on dataset subsets of different sizes for the image datasets using gradient few-shot learning. The reported difference is in comparison to using sample selection from the full dataset.}
    \label{fig:image-dataset_size_change}
\end{figure*}

In this section, we analyse how the benefit of sample selection strategies changes as we decrease the size of the dataset available. For this ablation study, we apply the \textit{ACSESS} method and \textit{Random} selection to select 5 samples per class from dataset subsets of different sizes and compare their benefit to the selection from the full dataset. Each dataset is subsampled into multiple sizes, starting with 5 samples per class up to the full dataset. Each subsampling is repeated 10 times and an average result over these repeats are reported. The full-dataset setting represents the selection from 200 samples per class for each datasets (for datasets with a lower  number of samples per class, such as HUM\_ACT.ACT\_410, MCR.PRT or MNF.TEX\_DTD, we observe a constant line after the full dataset was achieved). In all cases we perform selection of 5 shots per class using the ACSESS method and Random selection. Results aggregated over all datasets are in Figure~\ref{fig:dataset_size_change} and for individual datasets for the image datasets are included in Figure~\ref{fig:image-dataset_size_change}, for the text datasets using the gradient few-shot learning in Figure~\ref{fig:text-dataset_size_change} and for text datasets using the in-context learning in Figure~\ref{fig:icl-dataset_size_change}.

\textbf{Sample selection is beneficial even when using dataset with low number of samples.} Using only 25\% (50 samples per class) for in-context learning or 10\% of the dataset (20 samples per class) for gradient few-shot learning, sample selection can still identify samples that lead to similar performance as is achieved when using the full dataset. Decreasing the size further, the benefit of sample selection starts to decrease as well. Using only 10 samples per class, the benefit decreases by approximately $20-40\%$ (e.g., from performance increase of $2$ percentage points to only $1.3$), and then quickly deteriorates to the \textit{Random} selection (as the number of samples to choose from is too low).

For majority of the combinations of datasets and approaches, we can observe that sample selection can achieve similar performance using only a fraction of the dataset. For the gradient few-shot learning, this performance is achieved using only 10\% of the dataset (20 sample per class), while for in-context learning, this performance is achieved using 25\% of the dataset (50 samples per class). However, we observe that this strongly depends on the dataset and the model/approach used. For example on the ATIS dataset and in-context learning, using only 10 samples per class is enough to achieve the same performance as when selecting from the full dataset, while on the SNIPS dataset the number of samples per class increase up to 100. Similarly, on the ATIS dataset with gradient few-shot learning, the MAML a Few-Shot Fine-Tuning approaches achieve the similar performance using only 15-20 samples per class, while Prototypical Networks require the full dataset. Curiously, in specific cases, we can observe that running sample selection on subset of the dataset can lead to better performance than when performing selection from the full dataset (such as on the News Category dataset and gradient few-shot learning).

At the same time, we observe no change in impact of random selection from datasets of different sizes. In this case, the performance remains more or less constant (and often significantly below selection using the ACSESS method) throughout all the sizes -- only difference in performance stems from the random subsampling. Finally, using only 10 samples per class, we can observer that sample selection using the ACSESS method can lead to quite big performance improvement over the random selection. Therefore we can conclude that the sample selection is beneficial even when using a small fraction of the dataset size as the one used in the main experiments, although it will lead to a lowered benefit of the sample selection and its impact on the overall performance (which is still higher than when using random selection).
\section{Few-Shot Learning Problem Formulation}
\label{app:fsl-vs-icl}

In this section, we introduce the definition of the few-shot learning problem and the difference between gradient-based few-shot learning and few-shot learning via in-context learning. Note that in-context learning can be applied only to text datasets as it utilises pretrained large language models.

For the standard few-shot learning, we assume a larger collection of instances $D = {(x_i, y_i)}$, with larger number of classes. From this dataset, a separate task $D_T \subset D$ is created by randomly sampling few classes $N$. The task is characterised by a training set (also called support set, $S$) and test set (also called query set, $Q$). To create these sets, few samples ($k$) are sampled for each class. The few-shot learning algorithms then use the few samples from the support set $S_T$ to learn to solve the task $D_T$ quickly and are evaluated on this ability using the query set $Q_T$. This represents the $N$-way $k$-shot classification. The few-shot learning approaches mostly consist of two phases, training phase and testing phase. Often, these two phases work with separate sets of $D_T$, i.e., the classes appearing as part of tasks in the testing phase do not appear during training. As each task works only with subset of classes, the evaluation is often done on many such tasks and the average performance from these tasks is reported.

The gradient-based few-shot learning approaches follow episodic training similar to typical supervised training to gather knowledge and experience and learn to quickly adapt when encountering new task. Using the support set, the model is adapted to the task using gradient descent based approaches and then evaluated on the query set. Based on the evaluation, some kind of knowledge on how to better adapt is saved in part of the model, while the rest of the model is reset. This is done for a large number of episodes (often called meta-training), with the model accumulating the knowledge. After enough episodes are run, the model is evaluated (in a meta-testing phase) using the exact same setup.

On the other hand, no training phase is present in few-shot learning via in-context learning. Instead, a pretrained large language model is used without any additional parameter updates. For each new task, a task-specific text prompt is formed by concatenating task-specific instructions, the samples from the support set along with their labels, and a specific test sample. After being given this prompt, the large language model generates a prediction in the form of a word (or multiple words). These generated words are then mapped to the task labels, and a class is assigned. The process is repeated for each sample in the query set. For example, in the case of binary classification, the task-specific instruction may be "Determine sentiment of following sentence", and the concatenated samples from the support set may be formatted as "{sentence}, sentiment: positive/negative", with the word "positive" mapping to the class representing positive sentiment and "negative" to the class representing negative sentiment. As such, the only supervision for the large language model is from the instruction and the concatenation of support samples. Therefore, the model is limited by its maximum input length. At the same time, this approach was found to be sensitive to the selected samples in the support set and their order~\citep{zhang-etal-2022-active, pecher2023effects, chang-jia-2023-data}.

\end{document}